\documentclass{article} 
\usepackage{iclr2027_conference,times}

\usepackage{amsmath,amsfonts,bm}

\def\eqref#1{equation~\ref{#1}}

\def\1{\bm{1}}

\DeclareMathAlphabet{\mathsfit}{\encodingdefault}{\sfdefault}{m}{sl}
\SetMathAlphabet{\mathsfit}{bold}{\encodingdefault}{\sfdefault}{bx}{n}

\usepackage{hyperref}
\usepackage{url}
\usepackage{booktabs}
\usepackage{algorithm}
\usepackage{algorithmic}

\title{Some Memories Must Fade: Persistence and Decay of User Memories in Model Parameters}

\iclrfinalcopy
\usepackage{authblk}
\author[ ]{\textbf{Sanidhya Vijayvargiya}}
\author[ ]{\textbf{Rahul Lokesh}}
\affil[ ]{Samsung Research America}

\newcommand{\new}{\marginpar{NEW}}
\usepackage{pifont}
\usepackage{graphicx}
\usepackage{wasysym}
\usepackage{wrapfig}
\usepackage{multirow}
\usepackage{enumitem}
\usepackage{xcolor}
\usepackage{tcolorbox}
\usepackage{subcaption}
\usepackage{tikz}
\usetikzlibrary{positioning, arrows.meta, shapes.geometric, backgrounds}

\begin{document}

\newif\ifcamera
\cameratrue   
\ifcamera
  \long\def\new#1{#1}
  \newcommand{\newcolor}{}
\else
  \long\def\new#1{{\color{blue}#1}}
  \newcommand{\newcolor}{\color{blue}}
\fi

\maketitle

\vspace{-3mm}
\begin{abstract}
\vspace{-2mm}
Parametric memory allows Large Language Models (LLMs) to internalize knowledge directly into their weights, offering a potentially powerful alternative to retrieval-based memory (RAG). However, existing approaches to parametric memory are difficult to deploy as they largely mirror the Continual Learning (CL) objective of retaining information indefinitely. This is neither sustainable nor sufficient for a memory system, which must govern knowledge across its lifecycle by retaining stable facts, reinforcing recurring observations, updating changed beliefs, and allowing transient information to fade. We propose a system for parametric memory management consisting of a memory substrate (handling information storage and forgetting) and a learned policy (what to remember, how deeply, and for how long). Our substrate is a fixed-capacity LoRA adapter that stores memories in sparse $k$-hot combinations of adapter ranks to limit interference. This allows each rank and its stored information to maintain an independently controlled temporal decay, enabling outdated information to fade while long-term facts persist. On this substrate, we train a lightweight 2M-parameter controller that governs each incoming user event to ensure the memory system reflects the user's true, updated belief state. To evaluate how memory systems manage user memory, we introduce a controlled dataset of long-horizon user trajectories with varying persistence of facts based on changing belief states and user-behavioral shifts, with probe queries that test recall while information is valid and abstention after it expires. Our policy learns to effectively manage user memory over its entire lifecycle, balancing recall and abstention to achieve 86\% of oracle performance, unlike CL and RAG baselines that primarily default to one of the two behaviors. By providing the first system to manage memory directly in model weights, this work establishes a foundation for long-horizon parametric memory.
\end{abstract}

\vspace{-6mm}
\section{Introduction}
\vspace{-2mm}
Large Language Models (LLMs) increasingly operate over long horizons as agents and personalized assistants, necessitating memory systems to match evolving user preferences~\citep{ding2026alwaysonagentsasurveypersistentmemory,liu2025surveypersonalizedlargelanguage}. Currently, retrieval-based memory is widely adopted for its simplicity---storing information externally and injecting it into the context when relevant~\citep{lewis2021retrievalaugmentedgenerationknowledgeintensivenlp,kang2025memoryosaiagent}. Parametric memory instead encodes information directly into the model's weights, allowing LLMs to leverage memory implicitly without cluttering the prompt context~\citep{wang2025selfupdatablelargelanguagemodels,behrouz2024titanslearningmemorizetest,su2025parametricretrievalaugmentedgeneration}. However, this approach remains difficult to deploy in practice due to its rigidity in adapting to the evolving nature of user memory. Current approaches largely inherit the objective of continual learning (CL), focusing on how to acquire new knowledge without forgetting~\citep{MCCLOSKEY1989109, Kirkpatrick_2017}. We demonstrate that this objective is insufficient for long-horizon assistants. Memory management instead asks which knowledge should be retained, reinforced, overwritten, or forgotten, taking downstream use into account. For instance, \textit{I had a bagel for breakfast} should not remain valid even the next day, while \textit{I work as a teacher} should likely be relevant information for years. External stores expose lifecycle operations directly through addition, update, and deletion, but model parameters do not~\citep{yan2026memoryr1enhancinglargelanguage,yu2026agenticmemorylearningunified}. Consequently, in our evaluation, no CL method is able to represent outdated information even though they store values well (0.82 retrieval with PCGrad~\citep{yu2020gradientsurgerymultitasklearning}).

We hypothesize that usable parametric memory requires separating the storage dynamics (how information is stored, and isolated in the substrate) from its policy (what to store, how deeply, and for how long). For the substrate dynamics, we design a fixed-size LoRA substrate~\citep{hu2021loralowrankadaptationlarge} where each fact is routed to a $k$-hot combination of adapter ranks~\citep{zou2025flyloraboostingtaskdecoupling}. This bounds gradient interference between unrelated facts, while an independent temporal decay rate on each rank allows stale information to fade~\citep{ebbinghaus1885memory} without explicit unlearning. For the policy, we train a lightweight 2M-parameter controller that decides, per incoming event, whether to write, at what gradient depth, and with what lifespan. Stable information receives deep durable writes, repeated routines see accumulating writes, while transient events see short or no lifespans.

Evaluating these lifecycle decisions requires ground truth that existing memory benchmarks do not provide at the event level. We therefore construct $909$ controlled, long-horizon user trajectories with lifecycle annotations capturing validity, persistence, reinforcement, and contradiction, along with programmatically injected queries that probe per-event management decisions. On this testbed, our learned management system reaches $0.56$ class-balanced accuracy, compared with $0.42$ for the strongest CL baseline (PCGrad) and $0.50$ for a compute-matched retrieval-based memory manager, recovering 86\% of oracle performance. We study two questions: (1) Can a fixed-capacity parametric substrate support reliable information persistence and decay with varying per-fact dynamics? (2) Given such a substrate, can a policy infer from online observations how each fact should be managed? Our contributions are:
\vspace{-2mm}
\begin{itemize}[leftmargin=*, itemsep=0pt]
    \item \textbf{Task and controlled benchmark} (\S2):  We formalize parametric memory management around four requirements---valid retention, selective expiration, reinforcement and contradiction, and user-specific dynamics---and construct a dataset of long-horizon user trajectories with event-level lifecycle supervision and diagnostic probes.
    \item \textbf{Memory substrate} (\S3): We propose a fixed-capacity LoRA substrate combining sparse $k$-hot rank allocation with per-rank temporal decay, providing controllable mechanisms for selective persistence, reinforcement, overwrite, and calibrated expiration.
    \item \textbf{Learned management} (\S4): We define a $(\text{binary gate}, \text{depth}, \text{lifespan})$ action space over the substrate and train a 2M-parameter controller that recovers 86-92\% of oracle performance, demonstrating that effective lifecycle decisions can be inferred from the available online signals.
\end{itemize}

\vspace{-2mm}
\section{Task: Parametric Memory Management}
\vspace{-2mm}
\subsection{Memory Management Requirements}

We study parametric memory management for lifelong user-facing agents, where rather than preserving all written information, the system must manage each fact's lifecycle as user knowledge evolves. We identify four requirements:

\begin{enumerate}[leftmargin=*, itemsep=0pt]
\vspace{-2mm}
\item \textbf{Valid retention:} Recall stored information while it remains valid.
\item \textbf{Selective expiration:} Once information is no longer valid, abstain.
\item \textbf{Reinforcement and contradiction:} Repeated observations should strengthen an existing belief, while contradictory observations should weaken or overwrite it.
\item \textbf{User-specific dynamics:} Persistence and decay must be inferred from individual history, as the same type of information can be long-term for one user but transient for another.
\vspace{-2mm}
\end{enumerate}

Success requires controlling when and how individual memories persist, accumulate, change, or expire. Existing long-term memory benchmarks evaluate several related capabilities, including recall and belief revision, but none evaluate these lifecycle requirements at the event level (Table~\ref{tab:benchmark-comparison}).
\vspace{-1mm}
\subsection{Controlled Long-Horizon Trajectories}
\label{subsec:22}
\vspace{-1mm}
Evaluating these requirements requires ground truth for each fact at different stages in its lifecycle. Real long-term user histories generally do not provide this supervision. Privacy constraints limit collection, management events such as contradictions and temporary shifts are sparse, and the true validity of individual memories is generally unknown even when long interaction histories are available~\citep{uddin2026recallforgettingbenchmarkinglongterm}. 

We therefore generate a benchmark of 909 controlled, multi-year user trajectories (spanning 700–1,500 days) with dense event-level supervision from Gemini 3.7 Flash. Each trajectory is constructed via a five-stage pipeline: (1) sampling a behavioral motif (e.g., frequent routine changes) and a persona, (2) generating a multi-year backstory, (3) converting the backstory into a chronological event stream, (4) masking entities and injecting ground-truth labels for lifespan and stability, and (5) validating chronological consistency.

\begin{wrapfigure}{r}{0.5\textwidth}
\vspace{-7pt}
\centering
\includegraphics[width=\linewidth]{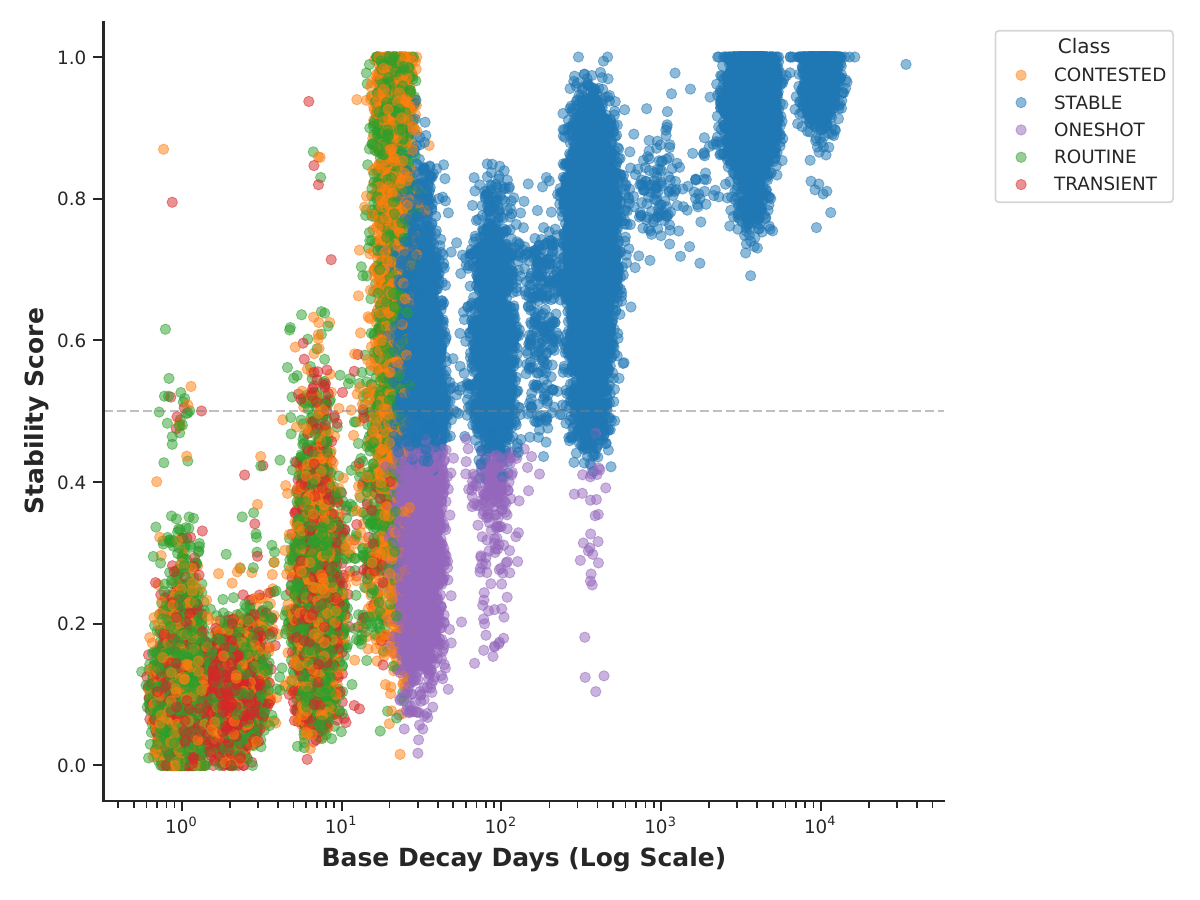}
\vspace{-2mm}
\caption{Distribution of the five memory classes by ground-truth stability score and validity horizon.}
\label{fig:data}
\vspace{-14pt}
\end{wrapfigure}

To prevent the model from relying on pretrained world knowledge, we replace concrete entities with typed placeholders inside fixed templates (e.g., replacing "apple" with \texttt{FRUIT\_4} in the template \textit{My favorite fruit is [MASK]}). Both writes and queries use this templated form as input \new{in our main experiments} and the model must output the valid value that should be in place of [MASK] or abstain if expired. By stripping away real-world priors, we force the controller to learn generalizable patterns that are not entity-specific or data generation artifacts (not memorizing typical lifespans for entities). The event-level annotations support both evaluation and training. They provide dense supervision over memory validity, lifespans, reinforcement, and contradiction, enabling supervised or imitation learning, dense reward construction, and end-to-end reinforcement learning for memory management (Appendix~\ref{app:dataset}).\footnote{We also provide naturalistic, non-templated versions of all inputs for evaluating retrieval-based methods. \new{\S\ref{sec:nl} runs the full learned system on these natural-language (NL) inputs.}}

\vspace{-2mm}
\subsection{Memory Dynamics}
\vspace{-1mm}
\label{sec:pattern_taxonomy}

\begin{table}[t]
\centering
\small
\setlength{\tabcolsep}{4pt}

\newcommand{\rot}[1]{\rotatebox{55}{#1}}
\newcommand{\yes}{\ding{51}}
\newcommand{\no}{\ding{55}}
\newcommand{\half}{\LEFTcircle}

\begin{tabular}{lccccc}
\toprule
& Belief $\Delta$
& Expiry
& Reinforcement
& User-dynamics
& Lifecycle GT \\
\midrule
ForgetBench~\citep{gu2026forgetbenchbenchmarkingforgettingdynamics}
    & \half & \no & \no & \no & \no \\
LoCoMo~\citep{maharana2024evaluatinglongtermconversationalmemory}
    & \half & \no & \no & \no & \no \\
PersonaMem~\citep{jiang2025knowmerespondme}
    & \yes & \no & \no & \half & \half \\
MemoryAgentBench~\citep{hu2026evaluatingmemoryllmagents}
    & \yes & \no & \no & \no & \no \\
LongMemEval~\citep{wu2025longmemevalbenchmarkingchatassistants}
    & \yes & \no & \no & \no & \half \\
Memora~\citep{uddin2026recallforgettingbenchmarkinglongterm}
    & \yes & \no & \half & \half & \yes \\
\midrule
\textbf{Ours}
    & \yes & \yes & \yes & \yes & \yes \\
\bottomrule
\end{tabular}

\vspace{-1mm}
\caption{Coverage of memory-management properties (\yes~full, \half~partial, \no~absent). Ours evaluates belief revision, time-based expiry, reinforcement, and user-conditioned lifespans while providing dense lifecycle supervision. Further details are in Appendix~\ref{app:benchmark_comparison}.}
\label{tab:benchmark-comparison}
\vspace{-5mm}
\end{table}

The benchmark requirements are informed by memory behaviors emphasized across prior work on long-term memory and personalization~\citep{maharana2024evaluatinglongtermconversationalmemory,wu2025longmemevalbenchmarkingchatassistants,jiang2025knowmerespondme,uddin2026recallforgettingbenchmarkinglongterm,mcclelland1995why,park2024memoriaresolvingfatefulforgetting}. Following generation, the facts are grouped into five management classes based on their annotated validity horizon, stability, reinforcement history, and contradictions:

\vspace{-2mm}
\begin{enumerate}[leftmargin=*, itemsep=0pt]
\item \textbf{STABLE (26\%):} Long-lived (90--3650 days), typically introduced once (e.g., profession). Requires durable retention over long horizons.
\item \textbf{ROUTINE (42\%):} Short-lived (2--21 days) observations that repeatedly recur with the same value. Requires evidence to accumulate rather than treating each observation independently.
\item \textbf{CONTESTED (20\%):} Information whose value is revised over time (e.g., current project). Requires updating the belief state to the latest value and superseding previous state.
\item \textbf{TRANSIENT (7\%):} Very short-lived (1--10 days), unrepeated information (e.g., today's lunch). Requires expiration and eventual abstention even without a subsequent contradiction.
\item \textbf{ONESHOT (4\%):} Unrepeated information with an intermediate validity horizon (15--90 days). Requires temporary retention without either immediate loss or indefinite lifespans.
\end{enumerate}
\vspace{-2mm}

\vspace{-1mm}
\subsection{Query Evaluation}
\vspace{-1mm}
We evaluate management using queries placed at controlled points in each fact's lifecycle. A \emph{retrieval query (VAL)} occurs while the fact is valid and is scored on recovering the current value. An \emph{abstention query (ABS)} occurs after expiry and is scored on answering \textit{Unknown}. For CONTESTED facts, queries target the currently valid value. Repeated ROUTINE events test whether consistent observations accumulate into stronger retention. From Figure~\ref{fig:data}, we can see that some classes have distinctive ground-truth patterns (STABLE vs TRANSIENT) but others can be overlapping (ROUTINE vs CONTESTED vs TRANSIENT). A valid management policy must therefore identify the user-level signals that segregate overlapping classes. Per-fact query budgets are reweighted to ensure that important but less frequent behaviors, such as expiration, receive sufficient evaluation mass despite write imbalances (Appendix~\ref{app:dataset}). The resulting query distribution is therefore designed as a controlled stress test of memory-management capabilities rather than an estimate of natural query frequency. Each class isolates a distinct challenge: long-horizon retention (STABLE), reinforcement accumulation (ROUTINE), belief revision (CONTESTED), principled expiration (TRANSIENT), and intermediate-horizon retention (ONESHOT). We report class-balanced (CB) accuracy so that performance is not dominated by the most frequent lifecycle class.

\vspace{-2mm}
\section{Memory Substrate}
\label{sec:substrate}
\vspace{-1mm}
\begin{figure}[t]
\vspace{-6mm}
 \centering
 \includegraphics[width=\textwidth]{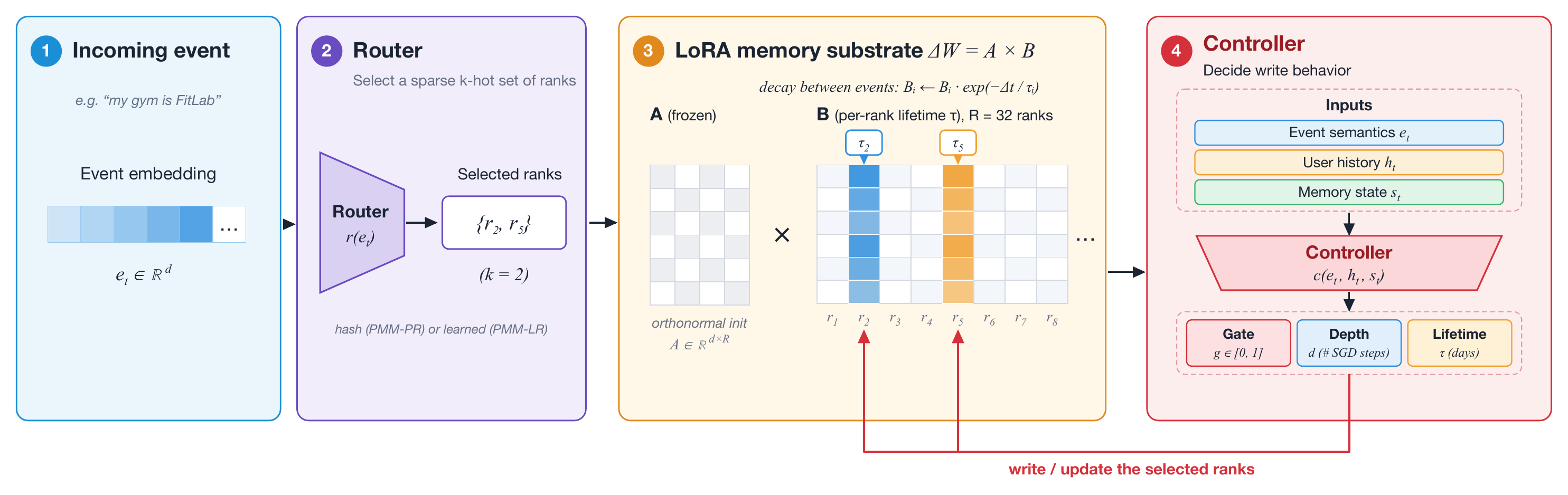}
 \vspace{-8mm}
 \caption{\textbf{Overview of the parametric memory management architecture.} An incoming user memory is mapped by the router to its storage in the LoRA adapter, consisting of a $k$-hot set of ranks in the up-projection (B) matrix. Based on the available online features including features from these ranks, the controller decides whether and how the information should be memorized.}
 \label{fig:main}
 \vspace{-15pt}
\end{figure}
Prior work shows that LoRA adapters can serve as compact parametric stores for information~\citep{xu2026loraremembersparametricmemory,back2026understandingloraknowledgememory}. Standard LoRA, however, is not well suited to lifelong memory management as consecutive writes update the same shared parameter space, causing interference between unrelated memories, while information written once has no mechanism to expire selectively~\citep{yang2024parametercollisionhinderingcontinual}. We therefore design a fixed-capacity substrate on a Qwen 3 1.7B backbone with three properties required for parametric memory management in the attached LoRA adapter: \textbf{isolated writes}, \textbf{controllable lifespans}, and \textbf{calibrated expiration}. Together, these provide the memory dynamics over which the management policy in \S\ref{sec:controller} operates (Figure~\ref{fig:main}). 
\vspace{-1mm}
\subsection{Sparse Combinatorial Memory Locations}
\label{sec:khot}
\vspace{-1mm}
We treat the $R$ rank columns of the adapter's $B$ matrix as an addressable memory bank. Each write is routed to a $k$-hot subset of columns while all others remain frozen and retrieval loads the same subset~\citep{zou2025flyloraboostingtaskdecoupling}. The $A$ matrix is orthonormally initialized and frozen throughout, as freezing $A$ yields a more predictable, approximately proportional relationship between write depth and memorization strength~\citep{zhang2026lorafaefficienteffectivelow}. With $R = 32$ and $k = 2$ as our main configuration, this yields $\binom{32}{2} = 496$ addressable locations from $32$ shared basis columns. Larger $k$ raises recall at the expense of abstention (Appendix~\ref{app:sweep}). This value-abstain trade-off dictates the limits of the management policies analyzed in~\S\ref{sec:controller}.

This routing provides two core properties. First, \textbf{interference is bounded} as writes to disjoint masks share no gradient path, and contention scales with mask overlap $|m_1 \cap m_2|$ only. Second, \textbf{repeated writes accumulate} as five writes at depth $d = 20$ (20 gradient descent steps at fixed learning rate) achieve comparable retention to a single $d = 100$ write (Appendix~\ref{app:physics}), allowing policies to reinforce facts via incremental updates. A key component of this architecture is the router that assigns related facts and queries to the same k-hot ranks for reliable updates and retrieval. We evaluate two routers. Enabled by our templated setting, the \emph{perfect router} deterministically hashes templates to masks, upper-bounding routing. The \emph{learned router} assigns slots using cosine similarity on the fact's whitened embedding~\citep{su2021whiteningsentencerepresentationsbetter} with existing slots, allocating new masks when similarity is low (Appendix~\ref{app:substrate}). Routing is a separable, independently tuned module. By evaluating both perfect and learned variants, we decouple the impact of storage from management. 
\vspace{-1mm}
\subsection{Controllable Temporal Lifespans}
\label{sec:decay}
\vspace{-1mm}
To support selective expiration, each rank column $i$ carries an independent decay constant $\tau_i$. Between events separated by $\Delta t$ days, $B[:, i] \leftarrow B[:, i] \cdot \exp(-\Delta t / \tau_i)$. Each write overwrites $\tau_i$ for its columns, letting the policy assign different lifespan horizons to different information. We discretize $\tau$ into five temporal scales using empirical decay timelines, from short-lived to effectively permanent (values in \S\ref{sec:action_space}). Per-location decay is structurally necessary because stable and transient information coexist in the same fixed-capacity adapter. A single global rate either expires stable facts prematurely or preserves transient ones indefinitely (validated in \S\ref{sec:results}, where a global $\tau$ reduces abstention on all volatile classes). Independent decay makes forgetting a controllable operation where information fades and its capacity is reclaimed without an explicit unlearning step. However, when two facts are routed to overlapping columns, the older fact's decay rate is overwritten by the newer fact's decay rate. We analyze the impact of this shared-column limitation in \S\ref{sec:no_free_lunch}.

\vspace{-1mm}
\subsection{Calibrated Retrieval and Abstention}
\label{sec:skill}
A memory substrate must also represent absence. A weak or expired trace should produce abstention rather than a hallucinated value. Base models tend to fabricate values from weak parametric traces, and prior work treats this behavior as an intrinsic threshold of parametric memorization~\citep{xu2026loraremembersparametricmemory}. We show that abstention from weak traces is instead a resolvable, calibration problem.

A separate rank-$32$ \emph{retrieval adapter} operates jointly with the memory adapter during inference, translating memory traces into expected outputs. Instead of memorizing facts itself, the retrieval adapter is trained to read the underlying substrate's state using depth-conditioned labels. If a routed slot is empty, it is trained to output \texttt{Unknown}. If a fact was written to the memory adapter at depth $d$, the retrieval adapter's target is \texttt{Unknown} with label probability $p_{\text{unk}}(d)$ and the correct value otherwise (e.g. $p_{\text{unk}}(20)=0.9$ and $p_{\text{unk}}(150)=0$).

The shape of the monotonically decreasing curve $p_{\text{unk}}(d)$ (Appendix~\ref{app:skill}) governs behavior under weak writes. A curve that drops too soon may retrieve hallucinated values from weak traces, while one that drops too late may collapse all queries to abstention. This probabilistic mapping yields a graded storage mechanism, rather than a binary retrievable threshold, giving the controller meaningful control over retention strength via $d$. Crucially, because the retrieval adapter uses the state of the substrate to generate answers rather than the input, retrieval is separated from memory management. The same substrate can be paired with alternative adapters, such as one trained for free-form natural language instead of our templates, without altering the underlying mechanics. (\S\ref{sec:nl}, Appendix~\ref{app:nl_study}).

\vspace{-1mm}
\subsection{Why Continual Learning Is Insufficient}
\label{sec:cl_comparison}
\vspace{-1mm}

\begin{wraptable}{r}{0.48\linewidth} 
  \centering
  \small

  \vspace{4pt} 
  \begin{tabular}{lccc}
    \toprule
    Method & CB $\uparrow$ & VAL $\uparrow$ & ABS $\uparrow$ \\
    \midrule
    Naive LoRA & 0.23 & 0.51 & 0.00 \\
    EWC        & 0.22 & 0.50 & 0.00 \\
    PCGrad     & 0.42 & \textbf{0.82} & 0.00 \\
    \midrule
    Ours (PMM-PR)  & \textbf{0.60} & 0.68 & \textbf{0.59} \\
    \bottomrule
  \end{tabular}
  \vspace{-1mm}
  \caption{CL protects stored information but cannot represent selective expiration. Per-class breakdowns are in the Appendix~\ref{app:cl}.}
  \vspace{-2mm}
  \label{tab:cl_summary}
\end{wraptable}Do these properties emerge from applying standard continual learning (CL) to a shared LoRA store? 
We test EWC~\citep{Kirkpatrick_2017} and PCGrad with sequential fixed-depth LoRA writes, selecting $d=15$ (optimal selection from $d=10,15,20,40,60$). Both reduce interference by protecting previously written information but provide no mechanism for selective expiration. Consequently, every CL baseline achieves $0.00$ ABS despite PCGrad reaching $0.82$ VAL (Table~\ref{tab:cl_summary}). Adding uniform exponential decay changes accuracy by $<0.01$, since dense shared parameters cannot assign different lifespans to different memories. Our substrate resolves this structurally as specific ranks decay to physically represent forgetting, which trigger the retrieval adapter to abstain. Sparse locations, independent decay, and calibrated retrieval provide this control.

\vspace{-2mm}
\section{Learned Memory Management}
\label{sec:controller}
\vspace{-1mm}

The substrate defines the storage of parametric memory. Here, we investigate (1) what actions are required to manage this substrate, (2) whether the information available online is sufficient to infer effective management decisions, and (3) what trade-offs remain even under effective management due to the substrate's finite capacity.

\vspace{-1mm}
\subsection{Memory Management Actions}
\label{sec:action_space}
\vspace{-1mm}

For each incoming event, the controller emits an action $(g,d,\tau)$:
\vspace{-2mm}
\begin{itemize}[leftmargin=*, itemsep=0pt]
\item \textbf{Gate} ($g \in {0,1}$): whether to write the event at all. Declining leaves the substrate unchanged.
\item \textbf{Depth} ($d \in {0,20,30,40,60,80,110,150}$ SGD steps): how strongly to write. Higher $d$ yields stronger retention but a larger gradient footprint on shared mask columns.
\item \textbf{Lifespan} ($\tau \in {25,100,400,1600,20000}$): the decay timescale assigned to the routed columns, controlling how long the resulting trace persists.
\end{itemize}
\vspace{-2mm}

The three actions are predicted by independent discretized heads, allowing each to be conditioned and bias-initialized separately (\S\ref{sec:expert}). Discretization also constrains the policy search space and simplifies optimization. We find two design choices to be important. First, the action space distinguishes \emph{rejecting} information ($g=0$) from \emph{briefly retaining} it through a shallow write with short lifespan. Different volatile events may require either behavior, motivating independent gate and lifespan controls. Second, routing is separated from the management policy. The router in Section~\ref{sec:khot} determines which ranks an event is written in, while the controller determines how/whether it is stored through gate, depth, and lifespan. We evaluate both perfect and learned routing to separate allocation errors from management errors. 
\vspace{-1mm}
\subsection{State and Policy Architecture}
\label{sec:features}
\vspace{-1mm}

The 2M-parameter controller ($<0.4\%$ of the base model) conditions on three sources of online information: (1) the \textbf{incoming event} represented by its semantic embedding, (2) \textbf{trajectory history} represented by a GRU state~\citep{cho2014learningphraserepresentationsusing}, and (3) \textbf{substrate state}, consisting of per-rank ages (time since last write for each rank), per-rank norms (to capture stored information), and token-level prediction signals from a forward pass that indicate whether the incoming event reinforces or contradicts the existing memory. A shared neural trunk feeds independent decision heads for gate, depth, and lifespan (Appendix~\ref{app:architecture}). Persistence cannot be determined from event content alone. Because identical facts exhibit different stabilities across users, relying solely on semantics leads to a prediction ceiling of only $R^2=0.36$ (Appendix~\ref{app:stagea}). The controller addresses this ambiguity by grounding its decisions in the user's trajectory history. Finally, to stabilize early training, the action heads are initialized with empirical biases derived from an expert policy.
\vspace{-2mm}
\new{
\vspace{-1mm}
\subsection{Pipeline Workflow}
\label{sec:workflow}
\vspace{-1mm}
\begin{figure}[t]
\vspace{-6mm}
 \centering
 \includegraphics[width=\textwidth]{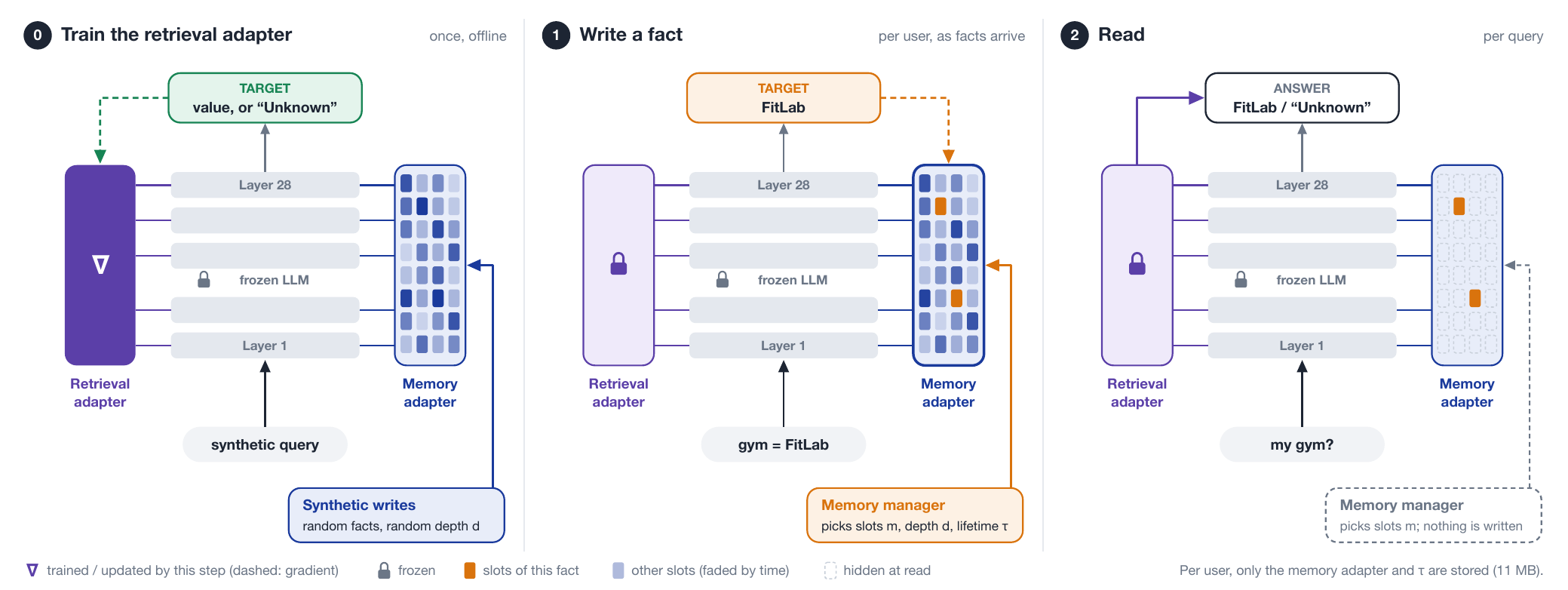}
 \vspace{-4mm}
 \caption{\new{End-to-end workflow of parametric memory management (PMM)}}
 \label{fig:workflow}
 \vspace{-15pt}
\end{figure}
\textbf{Components.} The system is composed of a frozen Qwen3-1.7B backbone, two additive LoRA adapters (Memory and Retrieval), and two management components (a router and a controller). Both adapters are attached to \texttt{q\_proj} and \texttt{v\_proj} in every layer, so a layer's output is $h = Wx + B_{\text{read}}A_{\text{read}}x + B_{\text{mem}}A_{\text{mem}}x$. The base model is never updated in the process (Figure~\ref{fig:workflow}).

\vspace{-1mm}
\textbf{Retrieval-adapter training (before deployment).} We simulate writes of random facts at random depths and train the retrieval adapter to output either the stored value or \texttt{Unknown}, according to the depth-conditioned label probability $p_{\text{unk}}(d)$ (\S\ref{sec:skill}). It is then frozen for all subsequent steps.
\vspace{-1mm}

\textbf{Write (per incoming fact).} The router maps the fact to an existing rank mask, or allocates a new one. The mask is specific to a fact family and shared across all layers (we use $k=2$ columns per family). The controller reads the fact embedding, the memory-adapter state, and the user history (through its GRU), and outputs whether to write the fact, a depth $d \in [20,150]$, and a lifetime $\tau \in [25, 20000]$ days. If written to the Memory adapter, the write is $d$ SGD steps on the selected columns of $B_{\text{mem}}$ only, and those columns take decay rate $\tau$. $A_{\text{mem}}$ is orthonormal and frozen throughout. The retrieval adapter is not active during writes.

\vspace{-1mm}
\textbf{Decay (between events).} Each column decays as $B_i \leftarrow B_i \exp(-\Delta t/\tau_i)$, making a fact expire.
\vspace{-1mm}

\textbf{Read (per query).} The router selects the query's mask, and ranks outside it are zeroed. The Memory and Retrieval adapters are both active, and the Retrieval adapter decides from the memory state whether to emit the stored value or \texttt{Unknown}. The controller is not used.
}
\vspace{-2mm}
\subsection{Learning from a Metadata Oracle}
\label{sec:expert}
\vspace{-1mm}
We train the controller by imitation from a manually configured \emph{causal metadata oracle}. The oracle observes the same online history as the controller together with ground-truth lifecycle metadata (stability, time to decay, and reinforcement status) and computes a target write depth based on a deficit. The deficit is the estimated depth required for a fact to remain retrievable over its target horizon, minus any depth already accumulated through prior writes, capped at 150. The oracle has no access to future events, but receives lifecycle metadata that the learned controller must infer from its observable state features. Crucially, this supervision adapts to the student's actual substrate state under DAgger~\citep{ross2011reductionimitationlearningstructured}. When the student deviates from the oracle, the target depth is recomputed from what the student has actually written rather than assuming ideal previous actions. Replacing this state-aware supervision with fixed target depths derived solely from ground-truth metadata reduces class-balanced accuracy from $0.56$ to $0.49$ (Appendix~\ref{app:expert}).
\vspace{-2mm}
\paragraph{Training.}
The training pipeline has three stages: (A) stability-head MSE training, (B) behavior cloning against the oracle via DAgger, and (C) optional PPO refinement~\citep{schulman2017proximalpolicyoptimizationalgorithms} (Appendix~\ref{app:phase2}). Stage B reaches near-oracle performance and is our primary configuration. We compare our parametric memory management (PMM) controllers trained with perfect routing (PMM-PR) and learned routing (PMM-LR) to separate the learned policy impact from allocation quality.
\vspace{-3mm}
\subsection{Results}
\label{sec:results}
\vspace{-2mm}
We evaluate on $15$ held-out user trajectories ($\sim600$ events per user) with sequential LoRA writes and queries. Results are reported per lifecycle class and per query type (VAL, ABS), aggregated using class-balanced (CB) accuracy. Table~\ref{tab:main_results} answers three questions. First, can conventional parametric memorization solve the lifecycle problem? No, CL baselines and unmanaged-write on our substrate retrieve values well but fail to expire outdated information. Second, is lifecycle management learnable? Yes, PMM-LR recovers 86\% of the metadata-oracle ceiling. Third, what limits performance? The fully learned PMM-LR reaches $0.56$ CB accuracy, while perfect routing raises performance to $0.60$ against a metadata-oracle ceiling of $0.65$. This hierarchy separates distinct sources of error. The PMM-LR-to-PMM-PR gap reflects semantic allocation, the PMM-PR-to-oracle gap primarily reflects lifecycle inference, and the oracle ceiling reflects limitations of the substrate itself, including shared-column interference between distinct memories and $\tau$ assignment (analyzed below). Improving routing can close the first gap, richer online management signals can close the second, and changes to allocation or capacity are required to raise the substrate ceiling. 

\begin{table}[t]
\centering
\scriptsize
\begin{tabular}{lcccccc}
\toprule
System & CB & STABLE & ROUTINE & CONTESTED & TRANSIENT & ONESHOT \\
\midrule
Metadata oracle & $0.65 \pm .03$ & $0.56 \pm .05$ & $0.72 \pm .06$ & $0.51 \pm .07$ & $0.77 \pm .05$ & $0.68 \pm .05$ \\
\midrule
PCGrad (best CL, $d=15$) & $0.42 \pm .02$ & $0.60 \pm .05$ & $0.66 \pm .05$ & $0.41 \pm .05$ & $0.13 \pm .05$ & $0.31 \pm .05$ \\
RAG ($1.7$B) & $0.50 \pm .02$ & $0.30 \pm .05$ & $0.42 \pm .05$ & $0.55 \pm .05$ & $0.75 \pm .05$ & $0.49 \pm .06$ \\
RAG-MM ($1.7$B) & $0.50 \pm .02$ & $0.41 \pm .05$ & $0.45 \pm .05$ & $0.48 \pm .06$ & $0.68 \pm .05$ & $0.47 \pm .05$ \\
RAG-MM ($122$B) & $0.53 \pm .02$ & $0.47 \pm .04$ & $0.69 \pm .05$ & $0.47 \pm .05$ & $0.33 \pm .06$ & $0.67 \pm .06$ \\
\new{RAG-MM ($1.7$B, trained)} & \new{$0.55 \pm .03$} & \new{$0.38 \pm .06$} & \new{$0.48 \pm .05$} & \new{$0.54 \pm .07$} & \new{$0.77 \pm .05$} & \new{$0.57 \pm .05$} \\
\midrule
Unmanaged $d=20$ & $0.45 \pm .02$ & $0.56 \pm .05$ & $0.63 \pm .05$ & $0.43 \pm .06$ & $0.26 \pm .05$ & $0.38 \pm .05$ \\
Unmanaged $d=80$ & $0.50 \pm .02$ & $0.65 \pm .05$ & $0.69 \pm .05$ & $0.51 \pm .06$ & $0.11 \pm .05$ & $0.54 \pm .05$ \\
\midrule
\textbf{PMM-LR} & $\mathbf{0.56 \pm .02}$ & $0.40 \pm .05$ & $0.76 \pm .05$ & $0.56 \pm .05$ & $0.44 \pm .06$ & $0.65 \pm .07$ \\
PMM-PR & $0.60 \pm .02$ & $0.39 \pm .05$ & $0.75 \pm .05$ & $0.57 \pm .06$ & $0.59 \pm .06$ & $0.68 \pm .06$ \\
\midrule
\new{Metadata oracle (NL)} & \new{$0.67 \pm .02$} & \new{$0.59 \pm .05$} & \new{$0.78 \pm .05$} & \new{$0.57 \pm .05$} & \new{$0.52 \pm .06$} & \new{$0.87 \pm .05$} \\
\new{PMM-PR (NL)} & \new{$0.60 \pm .03$} & \new{$0.45 \pm .05$} & \new{$0.78 \pm .05$} & \new{$0.58 \pm .05$} & \new{$0.47 \pm .06$} & \new{$0.73 \pm .07$} \\
\new{PMM-LR (NL)} & \new{$0.54 \pm .03$} & \new{$0.47 \pm .05$} & \new{$0.79 \pm .05$} & \new{$0.55 \pm .05$} & \new{$0.34 \pm .05$} & \new{$0.55 \pm .08$} \\
\midrule
No per-rank $\tau$ (PMM-PR) & $0.48 \pm .02$ & $0.57 \pm .05$ & $0.70 \pm .05$ & $0.44 \pm .05$ & $0.24 \pm .06$ & $0.47 \pm .06$ \\
\bottomrule
\end{tabular}
\vspace{-5pt}
\caption{Evaluation results on $n=15$ held-out users, $4584$ queries. RAG-MM refers to RAG with a write-time memory-management architecture (Appendix~\ref{app:rag}). Naive/EWC continual-learning variants ($\leq0.23$ CB), oracle-metadata RAG upper bounds, and no-$\tau$ unmanaged baselines are reported in Appendix~\ref{app:full_results}. NL refers to natural-language interfacing of the same trajectories. \new{$95\%$ percentile bootstrap CIs clustered by user (users resampled, 2{,}000 resamples).}}
\label{tab:main_results}
\vspace{-6mm}
\end{table}

Both controllers outperform all CL and unmanaged-write baselines, as well as the retrieval-based memory managers tested at compute parity. PMM-LR reaches $0.56$ versus $0.50$ for compute-matched RAG-MM and also exceeds RAG-MM with a substantially larger 122B management model ($0.53$). A trained manager in RAG-MM matches performance, though it often struggles to synthesize correct answers even with improved management. Retrieval surpasses our controllers when its management pass is replaced with ground-truth lifecycle information ($0.64$ at $1.7$B and $0.82$ at $122$B \S\ref{sec:no_free_lunch}), highlighting the importance of the management signal independent of the underlying memory substrate. Once information is written, queries require no retrieval over an external history and no context growth with trajectory length.

\vspace{-2mm}
\subsection{Per-Class Trade-offs and Failure Modes}
\label{sec:no_free_lunch}
\vspace{-1mm}

\begin{wrapfigure}{r}{0.55\textwidth}
\centering
\includegraphics[width=\linewidth]{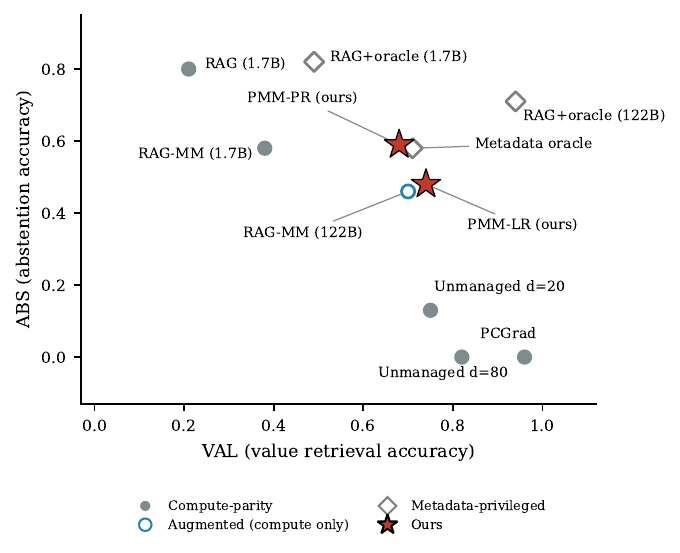}
\vspace{-20pt}
\caption{Pareto positions in value-retrieval (VAL) versus abstention (ABS) accuracy. Four regimes emerge: (1) continual learning and unmanaged writes at high VAL and near-zero ABS, (2) compute-parity retrieval at high ABS and low VAL, (3) augmented retrieval using greater compute or privileged lifecycle information reaching the top-right, and (4) our controllers occupying a balanced regime that no compute-parity baseline reaches. Breakdown in Table~\ref{tab:pareto}.}
\label{fig:valabs}
\vspace{-20pt}
\end{wrapfigure}\paragraph{Memorization without management fails at abstention.}
Fixed-depth writes demonstrate a clear trend. At $d=20$, VAL ranges from $0.60$-$0.90$ across classes while ABS remains only $0.01$-$0.22$. Increasing depth to $d=80$ raises VAL to $0.85$-$1.00$ while driving ABS to $\leq0.01$ (per-class VAL/ABS splits in Appendix~\ref{app:valabs}). Adding $\tau$ without event-specific management does not improve this trade-off (Appendix~\ref{app:blind_notau}).
\vspace{-1mm}
\paragraph{Retrieval-based memory also requires explicit management.}
Compute-matched simple RAG fails due to frequently defaulting to \texttt{Unknown} under ambiguous context (ABS $0.75$--$0.87$) while rarely committing to values (VAL $0.16$--$0.30$ on most classes). RAG-MM's write-time management pass---resolving reinforcement, contradiction, and expiry as events arrive (Appendix~\ref{app:rag})---shifts this Pareto position. Replacing the learned management pass with ground-truth lifecycle information raises RAG-MM to $0.64$ at $1.7$B and $0.82$ at $122$B (Appendix~\ref{app:full_results}), providing an upper bound when management is ideal and not inferred (Figure~\ref{fig:valabs}).

\paragraph{Management matters where events require different lifespans.}
Our controllers occupy a balanced regime that other compute-parity baseline do not reach. For TRANSIENT facts, no unmanaged fixed-depth policy exceeds $0.26$ accuracy because writing them strongly enough for immediate retrieval causes them to remain accessible after expiry. PMM-LR and PMM-PR reach $0.44$ and $0.59$, respectively, by assigning short lifespans to volatile information. Conversely, on STABLE facts, unmanaged $d=80$ reaches $0.65$ while PMM reaches $0.39$--$0.40$, pairing moderate VAL ($0.33$--$0.34$) with strong ABS ($0.78$--$0.81$). 
\vspace{-1mm}
\paragraph{The STABLE gap is due to $\tau$ overwrites on shared columns.}
Failures in retrieval of STABLE facts occur primarily at longer query horizons and under greater foreign-write interference, rather than from policy failures like declined writes or insufficient initial depth (Appendix~\ref{app:stable}). Because each write overwrites $\tau$ on the columns it touches, a STABLE fact can inherit the shorter timescale of a later volatile write sharing the same column over its long horizon. Preserving the maximum $\tau$ recovers ${\sim}13$ points of STABLE VAL but collapses ABS on volatile classes, reducing CB accuracy to $0.45$. This exposes a lifespan-compression trade-off in the shared substrate. Reinforcement partially mitigates the effect as STABLE facts with $3+$ writes achieve $0.68$ retrieval success as compared to $0.51$ for single-write facts. High STABLE retrieval performance can be achieved by routing them to non-overlapping ranks, at the cost of compression (Table~\ref{tab:dedicated}).

\begin{table}[t]
\newcolor
\centering
\small
\begin{tabular}{llcccc}
\toprule
Controller & Layout & CB & STABLE VAL & STABLE ABS & MB/user \\
\midrule
Oracle & 32 shared & .65 & .53 & .74 & 11 \\
Oracle & 64 shared & .70 & .73 & .58 & 22 \\
Oracle & 32 + 32 dedicated & .71 & .82 & .68 & 22 \\
\midrule
PMM-PR & 32 shared & .60 & .33 & .81 & 11 \\
PMM-PR & 32 + 32 dedicated & .63 & .70 & .67 & 22 \\
\midrule
PMM-LR & 32 shared & .56 & .34 & .78 & 11 \\
PMM-LR & 32 + 32, trained with routing & .59 & .75 & .27 & 22 \\
\bottomrule
\end{tabular}
\vspace{-1mm}
\caption{Dedicated columns for long-lived facts improve their retention. MB refers to size of adapter.}
\label{tab:dedicated}
\vspace{-4mm}
\end{table}
\vspace{-1mm}
\paragraph{Per-rank $\tau$ enables joint optimization.}
Removing per-rank lifespan control and forcing a global $\tau$ drops PMM-PR from $0.60$ to $0.48$. Without the ability to selectively expire information, the controller maintains retrieval only by sacrificing abstention. VAL remains $\geq0.79$ across volatile classes, while TRANSIENT ABS falls to $0.09$, ROUTINE ABS to $0.06$, and CONTESTED ABS to $0.04$. This establishes independent lifespan control as a necessary mechanism for supporting both retention and expiration in this substrate.

\begin{wraptable}{r}{0.5\linewidth}
\centering
\small
\begin{tabular}{l c l}
\toprule
Variant & CB acc & Notable effect \\
\midrule
PMM-PR (full) & $0.60$ & baseline \\
No history & $0.58$ & ONESHOT VAL $-24$ \\
No substrate & $0.58$ & ROUTINE ABS $-9$ \\
No priors & $0.52$ & overall $-8$pt \\
No semantic & $0.47$ & overall $-13$pt \\
\bottomrule
\end{tabular}
\caption{Ablations on controller input features.}
\label{tab:ablations_short}
\vspace{-3mm}
\end{wraptable}
\vspace{-1mm}
\paragraph{Ablations.}
Table~\ref{tab:ablations_short} reports input-feature ablations for PMM-PR. Removing semantic features or configured priors substantially reduces aggregate accuracy ($0.47$ and $0.52$, respectively), confirming that both provide non-redundant signals during training. History and substrate state are partially redundant because both reflect previous observations (through trajectory or through substrate state impact). However, class-specific failures highlight the distinct mechanistic role of each feature. Without trajectory history, ONESHOT VAL drops by 24 points because these first-time events lack a pre-existing trace to provide a redundant signal. Conversely, without substrate state, ROUTINE ABS drops by 9 points. Blind to how strongly a recurring fact is already embedded, the controller over-relies on historical frequency and over-writes the trace, preventing it from properly expiring. These results show that effective lifecycle management depends on event semantics, the user's trajectory, and the current state of memory. Full per-class results are provided in Appendix~\ref{app:ablations}.

\begingroup\newcolor
\vspace{-2mm}
\subsection{Natural Language Transfer}
\label{sec:nl}
\vspace{-1mm}

The experiments above use templated writes and queries to decouple management from memorization. Here we run the full learned system on the naturalistic version of the same trajectories, with the same architecture and training recipe. Natural-language writes behave differently (recall plateaus after about $60$ SGD steps), so we re-measured write strength and decay once, and refit the controller targets and the retrieval adapter on these measurements (Appendix~\ref{app:nl_study}). As seen in Table~\ref{tab:main_results}, every controller lands within $0.02$ CB of its templated counterpart.
\endgroup

\vspace{-2mm}
\section{Related Work}
\vspace{-1mm}
\subsection{Parametric Memory and Lifelong Knowledge Editing}
\vspace{-1mm}
A growing body of work stores information in model parameters or latent states rather than external text \citep{sun2025learninglearntesttime,ba2016usingfastweightsattend}. Recent systems divide into latent-state memory, maintaining pools of hidden-state memory tokens ~\citep{wang2024memoryllmselfupdatablelargelanguage, wang2025mextendingmemoryllmscalable}, and parametric memory, writing into weights through context distillation~\citep{wang2025selfupdatablelargelanguagemodels}, dedicated memory modules~\citep{berges2024memorylayersscale}, or test-time weight updates~\citep{behrouz2024titanslearningmemorizetest}. In parallel, lifelong knowledge editing develops efficient, localized modification of model knowledge ~\citep{meng2023locatingeditingfactualassociations,hartvigsen2023aginggracelifelongmodel,yu2023meloenhancingmodelediting}. Both lines study whether information can be written, retrieved, or updated while preserving unrelated knowledge. We instead study memory lifecycle management which explores how strongly information should be stored, how long it should persist, and when new observations should reinforce or replace existing beliefs. 

\vspace{-2mm}
\subsection{Sparse and Modular Parameter Updates}
\vspace{-1mm}
Sparse and modular weight updates reduce interference during updates. O-LoRA~\citep{wang2023orthogonalsubspacelearninglanguage} uses near-orthogonal task subspaces, N-LoRA~\citep{yang2024parametercollisionhinderingcontinual}, and FlyLoRA~\citep{zou2025flyloraboostingtaskdecoupling} allocate sparse adapter ranks to increase capacity and limit overlap. Our substrate similarly uses sparse combinations of adapter ranks, but as reusable memory locations rather than task-specific modules. Further, parametric overlap is not uniformly undesirable as unrelated writes should stay isolated, while overlap between related information supports reinforcement and overwrite. We further provide a controlled temporal decay for each rank, making persistence an explicit property of the memory rather than a byproduct of the memorization.
\vspace{-2mm}
\subsection{Memory Management and Continual Learning}
\vspace{-2mm}
To our knowledge, no prior work attempts to govern the parametric memory lifecycle---retention, expiration, reinforcement, and overwrite---as a learned management problem. Retrieval-based memory management systems organize external stores through explicit hierarchies and paging~\citep{packer2024memgptllmsoperatingsystems,kang2025memoryosaiagent}, learned memory operations~\citep{yu2026agenticmemorylearningunified,yan2026memoryr1enhancinglargelanguage}, or importance- and lifespan-based retention~\citep{park2024memoriaresolvingfatefulforgetting}, demonstrating that what is retained or forgotten can be explicitly managed. These managers, however, operate over explicit entries that can be directly added, updated, or deleted. Our controller instead governs the formation and lifespan of information encoded in parameters---deciding whether, how strongly, and for how long to write---so retention and forgetting emerge from the dynamics of the substrate itself.

Continual learning~\citep{Zeng_2019,yu2020gradientsurgerymultitasklearning,saha2021gradientprojectionmemorycontinual} addresses the complementary problem of acquiring new knowledge while minimizing interference with the old. Our experiments (\S\ref{sec:cl_comparison}) show memorization alone is insufficient for long-horizon memory, where some information must persist while other information must become inaccessible.

\vspace{-2mm}
\section{Conclusion}
\label{sec:conclusion}
\vspace{-2mm}

We present parametric memorization and parametric memory management as distinct problems, often conflated by prior work. Long-horizon memory requires deciding what to retain, reinforce, overwrite, or let expire. Existing parametric memory mechanisms do not explicitly expose or control these lifecycle decisions. Separating memory into its substrate dynamics---a fixed-capacity substrate with sparse, addressable locations and controllable temporal decay---and its policy---a lightweight controller governing the substrate through a defined action space---enables the system to balance retrieval of valid information against abstention on information that should no longer be accessible. Our controlled environment supports two central findings: (1) a fixed-capacity parametric substrate can support varying memory lifecycles, and (2) the decisions required to manage those lifecycles can be inferred from online observations.
\vspace{-2mm}
\paragraph{Limitations.}
\new{Moving to a new input format type requires re-measuring write strength and decay (\S\ref{sec:nl}, Appendix~\ref{app:nl_study}).} We also assume memory identification, determining what information constitutes a memory event, is given, and our controlled trajectories may not cover all real-world user dynamics. \new{Dedicated columns for long-lived facts increase per-user memory (\S\ref{sec:no_free_lunch}).}
\vspace{-2mm}
\paragraph{Future work.}
The learned-routing, controller, and oracle gaps motivate better semantic addressing, richer write-time signals, and improved capacity allocation. RL over long-term memory utility may further improve the controller, and different types of information like workflows can be stored in the substrate. 

Overall, our results suggest that reliable parametric memory requires progress beyond how models memorize. It requires controlling how information evolves after it has been written. By showing that a 2M-parameter policy operating on online signals recovers 86-92\% of a metadata-privileged oracle, we establish parametric memory management as a distinct and learnable problem and provide a foundation for lifecycle-aware memory in model parameters.

\bibliography{iclr2027_conference}
\bibliographystyle{iclr2027_conference}
\appendix
\section{Substrate Details}
\label{app:substrate}

This appendix expands the substrate measurements, routing mechanics, retrieval adapter training, and continual-learning comparison summarized in Section~\ref{sec:substrate}.

\subsection{Rank and $k$ Selection Data}
\label{app:sweep}

Table~\ref{tab:sweep_full} shows the full metadata oracle selection across
$R$ and $k$ with per-class VAL/ABS breakdown. All configurations use
real substrate with per-rank $\tau$ and each measurement is over $5$ held-out
users ($1452$ queries).

\begin{table}[h]
\centering
\scriptsize
\begin{tabular}{lccccccccccc}
\toprule
& \multicolumn{2}{c}{STABLE} & \multicolumn{2}{c}{ROUTINE} & \multicolumn{2}{c}{CONTESTED} & \multicolumn{2}{c}{TRANSIENT} & \multicolumn{2}{c}{ONESHOT} & \\
\cmidrule(lr){2-3} \cmidrule(lr){4-5} \cmidrule(lr){6-7} \cmidrule(lr){8-9} \cmidrule(lr){10-11}
Config & VAL & ABS & VAL & ABS & VAL & ABS & VAL & ABS & VAL & ABS & CB \\
\midrule
\multicolumn{12}{l}{\emph{Fixed $R = 32$, varying $k$}} \\
$k = 1$ & 0.58 & 0.70 & 0.33 & 0.83 & 0.42 & 0.53 & 0.26 & 0.89 & 0.62 & 0.77 & 0.61 \\
$k = 2$ & 0.64 & 0.70 & 0.76 & 0.74 & 0.85 & 0.27 & 0.46 & 0.80 & 0.97 & 0.46 & \textbf{0.69} \\
$k = 4$ & 0.57 & 0.75 & 0.97 & 0.58 & 1.00 & 0.15 & 0.92 & 0.41 & 0.97 & 0.62 & 0.66 \\
$k = 8$ & 0.49 & 0.75 & 0.95 & 0.40 & 1.00 & 0.09 & 0.97 & 0.21 & 0.92 & 0.48 & 0.57 \\
$k = 16$ & 0.36 & 0.75 & 0.90 & 0.21 & 0.98 & 0.06 & 0.97 & 0.09 & 0.75 & 0.33 & 0.48 \\
\midrule
\multicolumn{12}{l}{\emph{Fixed $k = 2$, varying $R$}} \\
$R = 16$ & 0.52 & 0.82 & 0.81 & 0.67 & 0.91 & 0.23 & 0.67 & 0.63 & 0.85 & 0.62 & 0.65 \\
$R = 32$ & 0.64 & 0.70 & 0.76 & 0.74 & 0.85 & 0.27 & 0.46 & 0.80 & 0.97 & 0.46 & 0.69 \\
$R = 64$ & 0.73 & 0.58 & 0.67 & 0.75 & 0.73 & 0.27 & 0.41 & 0.91 & 0.96 & 0.59 & \textbf{0.70} \\
$R = 128$ & 0.82 & 0.67 & 0.61 & 0.79 & 0.68 & 0.23 & 0.13 & 0.93 & 1.00 & 0.57 & 0.70 \\
\bottomrule
\end{tabular}
\caption{Full metadata oracle selection across substrate rank $R$ and mask
size $k$, with VAL/ABS split per class. The value-abstain trade-off is visible in both dimensions.
Increasing $k$ generally shifts volatile and revision-heavy classes toward higher VAL at the cost of ABS, while increasing $R$ redistributes
between classes without changing aggregate substantially above $R = 32$.}
\label{tab:sweep_full}
\end{table}

The per-class VAL/ABS decomposition surfaces two mechanisms:

\paragraph{Increasing $k$ shifts the value-abstention trade-off.}
CONTESTED VAL climbs from $0.85$ at $k = 2$ to $1.00$ at $k = 4$ and
above, but CONTESTED ABS falls from $0.27$ to $0.06$. The same pattern is observed
for TRANSIENT (VAL $0.46 \rightarrow 0.97$, ABS $0.80 \rightarrow
0.09$). Larger $k$ means each write touches more columns, so facts
are more likely to persist under any query (raising VAL) but also
less likely to expire correctly (lowering ABS). At $k = 16$, the
substrate is essentially always writing to overlapping columns and
per-fact $\tau$ has been almost entirely overwritten.

\paragraph{Increasing $R$ primarily improves STABLE.}
STABLE VAL climbs from $0.52$ at $R = 16$ to $0.82$ at $R = 128$, a
$30$-point improvement. This is exactly the mechanism we expect as
more rank columns means each STABLE fact is less likely to share
columns with subsequently-written short-$\tau$ facts. However,
ROUTINE and CONTESTED VAL both drop as $R$ grows because their
routing dispersion increases (the router allocates similar facts to
more distinct columns). The trade-off produces near-flat aggregate
accuracy above $R = 32$, at which point $R = 32$ is preferred for
its $2$--$4\times$ smaller inner-loop compute per write.

\paragraph{Dynamic router mechanics.} The dynamic router projects each fact's routing embedding through a learned whitening transform (fitted to dimension $256$), then compares it by cosine similarity against the accumulated signature of every currently allocated mask. A mask's signature is maintained as an $L_2$-normalized exponential moving average ($\alpha = 0.5$) of the embeddings previously routed there. If the highest similarity exceeds a threshold of $0.85$, the write is routed to that mask and its signature is updated, otherwise a new mask is allocated, with up to $\binom{R}{k}$ locations total.

\subsection{Substrate Dynamics Measurements}
\label{app:physics}

\paragraph{Retrieval after first write, full curve.} Table~\ref{tab:cold_curve_full} extends Figure/Table~2 (main text) to all tested depths.

\begin{table}[h]
\centering
\small
\begin{tabular}{lccccccccc}
\toprule
Depth $d$ & 10 & 15 & 20 & 30 & 40 & 60 & 80 & 110 & 150 \\
\midrule
Perfect router & 0.00 & 0.00 & 0.00 & 0.05 & 0.13 & 0.61 & 0.78 & 0.89 & 0.97 \\
Learnt router & 0.00 & 0.00 & 0.01 & 0.15 & 0.49 & 0.84 & 0.90 & 0.94 & 0.98 \\
\bottomrule
\end{tabular}
\caption{Retrieval after first write $P(\text{value})$ per depth.}
\label{tab:cold_curve_full}
\end{table}

\paragraph{Accumulation under repeated writes.} Table~\ref{tab:accumulation} supports the constructive-accumulation claim of \S\ref{sec:khot}. Five writes at $d = 20$ ($P \geq 0.93$) substantially exceed a single write at the same depth ($P \approx 0$ at $d=20$ per the table above), and approach the retention of a single deep write in the $d = 80$--$110$ range.

\begin{table}[h]
\centering
\small
\begin{tabular}{lcccc}
\toprule
& $n = 1$ & $n = 3$ & $n = 5$ & $n = 8$ \\
\midrule
$d = 20$ (perfect) & 0.00 & 0.29 & 0.93 & 0.97 \\
$d = 20$ (learnt) & 0.01 & 0.45 & 0.95 & 0.98 \\
\bottomrule
\end{tabular}
\caption{Cumulative retention $P(\text{value})$ after $n$ writes at
depth $d = 20$ with minimal intervening foreign traffic.}
\label{tab:accumulation}
\end{table}

\paragraph{Contradict-primed overwrite.} Table~\ref{tab:contradict} characterizes overwrite behavior (\S\ref{sec:cl_comparison}): a contradicting value written at $d \geq 40$ dominates retrieval under both routing configurations.

\begin{table}[h]
\centering
\small
\begin{tabular}{lcccc}
\toprule
Challenger depth & 20 & 40 & 80 & 110 \\
\midrule
$P(v_2)$ perfect router & 0.55 & 1.00 & 1.00 & 1.00 \\
$P(v_2)$ learnt router & 0.92 & 1.00 & 1.00 & 1.00 \\
\bottomrule
\end{tabular}
\caption{Contradict overwrite. Initial fact is written at $d = 40$ with value
$v_1$. The overwrite is a subsequent write with value $v_2$ at varying
depths. $P(v_2)$ is the retrieval probability of the overwrite.}
\label{tab:contradict}
\end{table}

\paragraph{Priming and decay validation.} To validate that per-rank exponential decay behaves as designed, we
run a simple test. A fact is written at depth $d$ with
$\tau = 1000$, decayed for $\Delta t$ days, and retrieved. Measured
retention matches predicted
$P(\text{cold}, d) \cdot \exp(-\Delta t / \tau)$ within measurement
noise for $d \in \{40, 80, 110\}$ and $\Delta t \in \{10, 100, 300\}$
days.

\paragraph{Slot norm regression details.} The slot norm regression takes as input a five-dimensional standardized
feature vector: self steps, foreign steps, $\log(1 + \text{self
steps})$, $\log(1 + \text{foreign steps})$, and time since last
write. Cross-validated $R^2$ is $0.87$ perfect routing and
$0.75$ on the learned router. The dynamic router's lower predictability
reflects that its slot signatures drift with fact allocation, adding
noise the regression cannot capture. This regression is what allows offline controller training to substitute predicted slot norm for a measured one (\S\ref{sec:features}), at a $15\times$ speedup.

\paragraph{Forward pass test.} Before each write, we compute $P(\text{value} \mid \text{current slot state}) -
P(\text{value} \mid \text{clean slot})$This provides a substrate-derived signal of whether the incoming value is already represented in the routed slot. The difference is positive for reinforcement events and near zero for first-write events. AUC against the ground-truth reinforcement labels is $0.99$
on both routers.

\begingroup\newcolor
\paragraph{Cost and latency.}
Table~\ref{tab:latency} compares PMM with RAG-MM, measured on $5$ users. Reads take a constant $94$\,ms regardless of how many facts are stored, versus $360$\,ms for RAG-MM, whose cost grows with the retrieved context. Writes are more expensive and depend on depth ($1.8$\,s at $d=20$ to $13.1$\,s at $d=150$, versus $2.4$\,s), but they can run asynchronously, off the query path. Over a full user trajectory, PMM costs $702$\,s against $617$\,s ($+14\%$), and is cheaper for $2$ of the $5$ users.
\endgroup

\begin{table}[h]
\newcolor
\centering
\small
\begin{tabular}{lcc}
\toprule
 & PMM & RAG-MM \\
\midrule
Read (per query) & 94 ms & 360 ms \\
Write (per fact) & 1.8 s ($d{=}20$) -- 13.1 s ($d{=}150$) & 2.4 s \\
Full user trajectory & 702 s & 617 s \\
\bottomrule
\end{tabular}
\caption{End-to-end timing ($5$ users).}
\label{tab:latency}
\end{table}

\subsection{Retrieval Adapter Training}
\label{app:skill}

We train a rank-$32$ retrieval adapter jointly loaded with the memory
adapter during both writes and reads. The training data is a mixed
recipe: some facts get shallow writes ($d \in \{10, 15, 20\}$), some
medium ($d \in \{30, 40\}$), some deep ($d \in \{60, 80\}$), some
very deep ($d \in \{100, 130\}$), and some slots are left empty.
For each written fact, we sample the target as \texttt{Unknown} with
probability $p_{\text{unk}}(d)$ or the correct value otherwise.

Table~\ref{tab:noise_curve} shows the final noise curve alongside the
implied target retrieval probability. The constraint $p_{\text{unk}}(d)
\leq 0.55$ prevents the retrieval from collapsing to always-abstain at
any depth.

\begin{table}[h]
\centering
\small
\begin{tabular}{lccccccc}
\toprule
Depth $d$ & 10 & 20 & 30 & 40 & 60 & 80 & 110+ \\
\midrule
$p_{\text{unk}}(d)$ & 0.55 & 0.38 & 0.30 & 0.22 & 0.12 & 0.05 & 0.00 \\
Implied $P(\text{value})$ & 0.10 & 0.20 & 0.32 & 0.45 & 0.70 & 0.85 & 0.95+ \\
\bottomrule
\end{tabular}
\caption{Retrieval adapter noise curve. Each depth targets a specific
$P(\text{value})$ during SFT.}
\label{tab:noise_curve}
\end{table}

\subsection{Continual Learning Baseline Details}
\label{app:cl}

We evaluate two continual-learning methods against our substrate,
all applied to LoRA writes at fixed depth $d = 15$ (chosen from $d \in \{15, 30, 60\}$ for each method). $d = 15$
maximized each method's aggregate accuracy. All CL baselines use the same base model and LoRA (rank $32$ on
$q_{\text{proj}}$, $v_{\text{proj}}$) but with fixed depth and no
per-rank $\tau$.

\paragraph{Naive fine-tuning.}
Direct backprop on the LoRA parameters at each write event, no
regularization. Serves as the floor baseline for LoRA-based
approaches.

\paragraph{EWC (Elastic Weight Consolidation).}
Adds a Fisher-weighted penalty term to write losses:
$\mathcal{L}_{\text{total}} = \mathcal{L}_{\text{write}} + \lambda
\sum_i F_i (\theta_i - \theta_i^\ast)^2$ where $F_i$ is the diagonal
Fisher information at parameter $i$ estimated from the calibration
data, $\theta_i^\ast$ is the pre-write parameter, and $\lambda =
1000$ (from~\citep{Kirkpatrick_2017}). Fisher is updated with exponential moving
average, $\alpha = 0.9$.

\paragraph{PCGrad (Projected Conflicting Gradients).}
Maintains a running memory of the $k = 5$ most recent write
gradients. Each new write gradient is projected to be orthogonal to
any stored gradient with which it has negative cosine similarity,
following~\citep{yu2020gradientsurgerymultitasklearning}. This is designed to prevent new writes
from destroying prior ones.

\paragraph{Retrieval adapter compatibility.}
CL baselines cannot use our retrieval adapter directly because the resoute
was trained assuming the $k$-hot mask retrieval. For fair comparison,
we retrained a compatible retrieval adapter for each CL baseline that
reads from the full LoRA weight rather than a masked subset. This ensures accuracy differences
reflect substrate design, not reading calibration mismatches.

\section{Learned Memory Management Details}
\label{app:policy}

This appendix expands the controller architecture, expert policy,
training details, per-class breakdowns, retrieval baseline
architecture, and negative results summarized in
Section~\ref{sec:controller}.

\subsection{Full VAL/ABS Breakdown}
\label{app:valabs}

Table~\ref{tab:pareto} shows the per-pattern accuracy split by target
type (VAL versus ABS) for the systems in Table~\ref{tab:main_results}
and Appendix~\ref{app:full_results}. 
\begin{table}[h]
\centering
\scriptsize
\begin{tabular}{lcccccccccc}
\toprule
& \multicolumn{2}{c}{STABLE} & \multicolumn{2}{c}{ROUTINE} & \multicolumn{2}{c}{CONTESTED} & \multicolumn{2}{c}{TRANSIENT} & \multicolumn{2}{c}{ONESHOT} \\
\cmidrule(lr){2-3} \cmidrule(lr){4-5} \cmidrule(lr){6-7} \cmidrule(lr){8-9} \cmidrule(lr){10-11}
System & VAL & ABS & VAL & ABS & VAL & ABS & VAL & ABS & VAL & ABS \\
\midrule
Metadata oracle & 0.53 & 0.74 & 0.75 & 0.65 & 0.84 & 0.24 & 0.52 & 0.81 & 0.93 & 0.46 \\
\midrule
Naive $d=15$ & 0.17 & 0.00 & 0.72 & 0.00 & 0.63 & 0.00 & 0.76 & 0.00 & 0.26 & 0.00 \\
EWC $d=15$ & 0.19 & 0.00 & 0.73 & 0.00 & 0.64 & 0.00 & 0.76 & 0.01 & 0.19 & 0.00 \\
PCGrad $d=15$ & 0.70 & 0.00 & 0.96 & 0.00 & 0.89 & 0.00 & 0.91 & 0.00 & 0.64 & 0.00 \\
\midrule
Unmanaged $d=20$ & 0.64 & 0.09 & 0.90 & 0.01 & 0.87 & 0.06 & 0.77 & 0.22 & 0.55 & 0.27 \\
Unmanaged $d=40$ & 0.81 & 0.00 & 0.99 & 0.00 & 0.99 & 0.01 & 0.96 & 0.02 & 0.92 & 0.03 \\
Unmanaged $d=80$ & 0.85 & 0.01 & 0.99 & 0.00 & 0.98 & 0.00 & 1.00 & 0.00 & 0.97 & 0.01 \\
Unmanaged $d=110$ & 0.83 & 0.00 & 1.00 & 0.00 & 1.00 & 0.00 & 1.00 & 0.00 & 0.97 & 0.00 \\
\midrule
RAG (1.7B) & 0.22 & 0.76 & 0.27 & 0.75 & 0.16 & 0.87 & 0.26 & 0.84 & 0.16 & 0.79 \\
RAG-MM (1.7B) & 0.39 & 0.53 & 0.45 & 0.46 & 0.38 & 0.56 & 0.36 & 0.74 & 0.30 & 0.63 \\
RAG-MM (122B) & 0.44 & 0.65 & 0.81 & 0.42 & 0.75 & 0.24 & 0.94 & 0.22 & 0.55 & 0.78 \\
\new{RAG-MM (1.7B, trained)} & \new{0.31} & \new{0.83} & \new{0.38} & \new{0.69} & \new{0.30} & \new{0.75} & \new{0.40} & \new{0.83} & \new{0.29} & \new{0.82} \\
RAG + oracle metadata (1.7B) & 0.39 & 0.78 & 0.55 & 0.65 & 0.50 & 0.75 & 0.60 & 0.90 & 0.41 & 1.00 \\
RAG + oracle metadata (122B) & 0.86 & 0.77 & 0.84 & 0.54 & 1.00 & 0.53 & 1.00 & 0.75 & 1.00 & 0.97 \\
\midrule
\textbf{PMM-PR (ours)} & 0.33 & 0.81 & 0.80 & 0.64 & 0.81 & 0.34 & 0.66 & 0.58 & 0.82 & 0.58 \\
\textbf{PMM-LR (ours)} & 0.34 & 0.78 & 0.88 & 0.47 & 0.93 & 0.21 & 0.83 & 0.35 & 0.71 & 0.60 \\
\midrule
No per-rank $\tau$ (PMM-PR) & 0.55 & 0.68 & 0.96 & 0.06 & 0.87 & 0.04 & 0.83 & 0.09 & 0.79 & 0.22 \\
\midrule
\new{Metadata oracle (NL)} & \new{0.55} & \new{0.85} & \new{0.86} & \new{0.58} & \new{0.95} & \new{0.22} & \new{0.90} & \new{0.46} & \new{0.93} & \new{0.81} \\
\new{PMM-PR (NL)} & \new{0.36} & \new{0.97} & \new{0.93} & \new{0.40} & \new{0.94} & \new{0.24} & \new{0.92} & \new{0.40} & \new{0.62} & \new{0.86} \\
\new{PMM-LR (NL)} & \new{0.40} & \new{0.92} & \new{0.94} & \new{0.40} & \new{0.96} & \new{0.17} & \new{0.97} & \new{0.24} & \new{0.45} & \new{0.65} \\
\bottomrule
\end{tabular}
\caption{Per-pattern accuracy split into VAL and ABS query types.
$n = 15$ held-out users.}
\label{tab:pareto}
\end{table}

\subsection{Unmanaged Baselines Without $\tau$}
\label{app:blind_notau}

For completeness, we ran the unmanaged baseline depth selections both with per-rank
$\tau$ (main table) and without $\tau$ (this table).

\begin{table}[h]
\centering
\small
\begin{tabular}{lcccccc}
\toprule
System & CB & STABLE & ROUTINE & CONTESTED & TRANSIENT & ONESHOT \\
\midrule
$d = 20$  & $0.45$ & $0.54$ & $0.62 $ & $0.41 $ & $0.30$ & $0.40$ \\
$d = 40$ & $0.49$ & $0.70$ & $0.68$ & $0.46$ & $0.15$ & $0.45$ \\
$d = 80$  & $0.50$ & $0.73$ & $0.68$ & $0.45$ & $0.15$ & $0.47$ \\
$d = 110$ & $0.49$ & $0.71$ & $0.69$ & $0.46$ & $0.15$ & $0.47$ \\
\bottomrule
\end{tabular}
\caption{Unmanaged baselines without $\tau$ are nearly identical to
their with-$\tau$ counterparts.}
\label{tab:blind_notau}
\end{table}

The near-identical aggregate numbers (all within $\pm 0.01$ CI
overlap) demonstrate an important point that $\tau$ alone is not
what enables our controller's performance. The controller's benefit
comes from combining per-fact $\tau$ policy with per-fact depth and
gate decisions.

\subsection{Retrieval Baseline Details}
\label{app:rag}

All retrieval baselines share three components: an external text store containing the naturalistic realization of each event, a BGE-small embedder (\(33\)M parameters) for indexing and query-time retrieval, and a generator. Because the data is an event stream, no text chunking is required and facts are embedded directly as individual, timestamped strings (e.g., \texttt{[Clock: 0045] I cook dinner}). At query time, the system retrieves the top $k=5$ most similar entries. The generator prompt explicitly instructs the model to read the retrieved timestamped entries and output \texttt{Unknown} if the memory does not clearly support an answer or appears outdated.

The variants differ only in how the store is maintained:

\begin{itemize}[leftmargin=*]
\item \textbf{RAG:} the standard baseline. Events are appended to the store as they arrive and no management is applied. Retrieval at query time returns the top-$k$ most similar entries.
\item \textbf{RAG-MM:} the store is actively managed at write time by an LLM management pass. As each event arrives, the pass resolves its relationship to existing entries---merging reinforcements of the same fact, superseding contradicted values, expiring entries past their inferred validity, and deduplicating.  It receives the new fact and retrieved candidates, and outputs a JSON decision to either ADD, UPDATE (soft-deleting the old entry), REINFORCE (merging and extending lifespan), or IGNORE. It also assigns an estimated decay in days (e.g., Transient: $7.0$, Routine: $30.0$, Stable: $3650.0$).
\item \textbf{RAG + oracle metadata:} identical to RAG-MM except the LLM management pass is replaced by the ground-truth lifecycle policy, using each event's generation-time metadata (validity horizon, contradiction and reinforcement status) to maintain the store perfectly. This removes lifecycle-management error while retaining the same retrieval and generation pipeline, providing an upper bound on retrieval-side memory management under privileged lifecycle information.
\end{itemize}

\paragraph{Generators.} Compute-matched rows (labeled $1.7$B) use
Qwen3-$1.7$B, the same base model as our substrate. Large-generator rows (labeled $122$B) use
Qwen~3.5 $122$B~A10B.
\subsection{Controller Architecture Details}
\label{app:architecture}

The learned management system is a 2M-parameter neural network consisting of two primary modules: a State Representation Engine (which maintains trajectory context and extracts semantic features) and the Action Policy (which predicts lifecycle management decisions). All internal representations use a hidden dimension of $d=256$.

\paragraph{1. State Representation Engine}
The representation engine processes the raw outputs of the base language model into dense, controller-ready features. Let $x_t \in \mathbb{R}^{2048}$ be the last-token embedding of the incoming event template from the base model.
\begin{enumerate}
    \item \textbf{Semantic Projection:} The raw embedding is normalized via a fitted input statistics layer and projected to a dense representation to decouple it from lexical noise: $z_t = \text{MLP}_{\text{sem}}(x_t) \in \mathbb{R}^{256}$.
    \item \textbf{Trajectory GRU:} To capture user-specific behavioral context, a Gated Recurrent Unit (GRU) maintains a hidden state $h_t \in \mathbb{R}^{256}$. At each event, the GRU consumes the dense embedding $z_t$ and the token-level forward pass result (probability of fact before writing) $p_{\text{hit}} \in [0,1]$ (indicating if the model already knows the fact): 
    $$h_t = \text{GRUCell}(x_{\text{in}} = [z_t \oplus p_{\text{hit}}], \, h_{t-1})$$
\end{enumerate}

\paragraph{2. Auxiliary State Heads}
Before computing actions, the network predicts the fact's structural properties using one auxiliary head attached to the representation engine. Because this property depend only on semantics (not the substrate), its input is detached from the computational graph to prevent auxiliary gradients from corrupting the shared representation trunk.
\begin{itemize}
    \item \textbf{Stability Head:} A scalar prediction of the fact's true stability score, optimized via Mean Squared Error (MSE): $\hat{s}_t = \sigma(\text{Linear}(z_t \oplus h_t)) \in (0, 1)$.
\end{itemize}

\paragraph{3. Action Policy and Shared Trunk}
The main controller determines the action tuple $(\text{gate}, \text{depth}, \tau)$ by observing both the semantic trajectory and the state of the underlying LoRA substrate. The input to the shared policy trunk is the concatenation of six feature families:
$$ s_{\text{policy}} = \Big[ z_t \oplus h_t \oplus A \oplus ||W_{\text{slot}}||_2 \oplus \hat{s}_t \oplus F_{\text{rank}} \oplus p_{\text{hit}} \Big] $$
where $A \in \mathbb{R}^R$ are the rank ages, $||W_{\text{slot}}||_2$ is the predicted or measured $L_2$ norm of the active LoRA mask, and $F_{\text{rank}}$ are the substrate interference features (foreign and self steps).

\paragraph{4. Decision Heads and Priors}
The policy state $s_{\text{policy}}$ is passed through independent linear heads to produce the final action logits. The model utilizes empirical priors to stabilize training and guide the depth decision:
\begin{itemize}
    \item \textbf{Gate Head ($g \in \{0, 1\}$):} Outputs a single logit $l_{\text{gate}}$. To prevent the gate from collapsing to $0$ early in training, the logit is initialized with an empirical bias $b_{\text{gate}}$ derived from the training set's marginal write rate: $P(\text{write}) = \sigma(l_{\text{gate}} + b_{\text{gate}})$.
    \item \textbf{Persistence Head ($\tau \in \{25, 100, 400, 1600, 20000\}$):} Outputs a 5-dimensional logit vector $l_{\tau}$, shifted by empirical class frequency biases to account for the heavy skew toward shorter decay horizons.
    \item \textbf{Depth Head ($d \in \mathcal{D}_{>0}$):} Outputs a 7-dimensional logit vector $l_{\text{depth}}$ over the available non-zero write depths $\{20, 30, 40, 60, 80, 110, 150\}$. The depth logits are directly augmented by the auxiliary head, acting as a learned prior.
    $$ l_{\text{depth}}^{\text{final}} = l_{\text{depth}} + b_{\text{empirical}} + \alpha_1 (\hat{s}_t V_{\text{stab}}) $$
    where $V_{\text{stab}}$ is a vector pushing high-stability predictions toward deeper bins.
\end{itemize}

\subsection{Training Dynamics and Loss Formulation}
\label{app:training_dynamics}

Training a shared representation trunk alongside multiple action heads presents significant optimization challenges. To stabilize Behavioral Cloning (Phase 1) and ensure smooth convergence, we employ three specific training techniques.

\paragraph{1. Ordinal Soft Targets for Depth Classification}
Standard cross-entropy loss heavily penalizes any deviation from the target class. However, write depth $d$ is an ordinal variable where adjacent bins represent similar retention strengths. To reflect this, the magnitude head is trained via Kullback-Leibler (KL) divergence against ordinal soft targets. If the expert deficit calculation targets depth $d^*$, the target distribution smears probability mass across adjacent available bins according to proximity, rather than placing a one-hot $1.0$ on the nearest bin. This ensures that near-miss predictions (e.g., predicting $d=80$ when the target is $110$) are penalized less than catastrophic under-writes.

\paragraph{2. DAgger Expert Annealing}
During Phase 1, we use DAgger~\citep{ross2011reductionimitationlearningstructured} to supervise the policy over the substrate. The probability $\beta$ of executing the expert's target action versus the student policy's sampled action follows a linear decay schedule. For the first $5\%$ of training (warmup), $\beta = 0.95$, ensuring the substrate remains sound while the trunk learns. Over the remainder of training, $\beta$ decays linearly to $0.30$, forcing the student to increasingly manage the consequences of its own imperfect writes.

\paragraph{Total Phase 1 Loss Formulation}
The complete multi-task objective during Behavioral Cloning is:
\begin{align*}
    \mathcal{L}_{\text{total}} = {} & w_{\text{ctrl}} \Big( \lambda_g \mathcal{L}_{\text{BCE}}(\text{gate}) + \lambda_d \mathcal{L}_{\text{KL}}(\text{depth}) + \lambda_\tau \mathcal{L}_{\text{CE}}(\tau) \Big) \\
    & + \lambda_s \mathcal{L}_{\text{MSE}}(\text{stability}) 
\end{align*}
where $w_{\text{ctrl}}$ is the warmup schedule scaler, and $\lambda$ are static hyperparameter weights balancing the relative entropy of the action spaces.

\subsection{Full Results: Additional Baselines}
\label{app:full_results}

Table~\ref{tab:full_results} gives the remaining systems omitted from
Table~\ref{tab:main_results} for space.

\begin{table}[h]
\centering
\small
\begin{tabular}{lcccccc}
\toprule
System & CB & STABLE & ROUTINE & CONTESTED & TRANSIENT & ONESHOT \\
\midrule
Naive & $0.23$ & $0.14 \pm .05$ & $0.50 \pm .05$ & $0.29 \pm .05$ & $0.11 \pm .04$ & $0.12 \pm .04$ \\
EWC & $0.22$ & $0.16 \pm .04$ & $0.50 \pm .05$ & $0.29 \pm .05$ & $0.12 \pm .05$ & $0.09 \pm .04$ \\
\midrule
Oracle RAG (1.7B) & $0.64$ & $0.45 \pm .05$ & $0.58 \pm .05$ & $0.63 \pm .06$ & $0.86 \pm .06$ & $0.71 \pm .05$ \\
Oracle RAG (122B) & $0.82$ & $0.84 \pm .04$ & $0.75 \pm .05$ & $0.75 \pm .05$ & $0.79 \pm .05$ & $0.99 \pm .04$ \\
\bottomrule
\end{tabular}
\caption{Additional baselines omitted from Table~\ref{tab:main_results}
for space: weaker continual-learning variants and retrieval
given ground-truth lifecycle metadata in place of a learned or
LLM-managed store.}
\label{tab:full_results}
\end{table}

\subsection{Expert Policy Details}
\label{app:expert}

The deficit-based expert (\texttt{capped\_deficit})
computes, for each write event, a scalar deficit and returns an
action tuple. 

\begin{algorithm}[h]
\caption{Deficit-based expert policy}
\begin{algorithmic}[1]
\STATE \textbf{Inputs:} stability score $s$, \texttt{is\_reinforcement}
flag $r$, base decay $b$, self-steps on mask $S$, foreign-steps on
mask $F$, depth bins $\{d_i\}$, foreign threshold $F^\ast$, tolerance
$\epsilon$, per-event cap $C = 150$.
\STATE Compute target retention depth $D^\ast = D^\ast(s, b, r)$
based on substrate curves.
\STATE Compute effective accumulated depth $D_{\text{acc}} = S -
\gamma F$ with foreign discount $\gamma = \min(1, F / F^\ast)$.
\STATE Compute deficit $\Delta = D^\ast - D_{\text{acc}}$.
\IF{$\Delta \leq \epsilon$}
    \STATE \textbf{return} $(g = 0, d = 0, \tau = 0)$ \COMMENT{decline}
\ENDIF
\STATE Set write depth $d^\ast = \min(\Delta, C)$.
\STATE Snap $d^\ast$ to nearest available bin.
\STATE Set $\tau^\ast = \arg\min_b |\log \tau_b - \log(24.5 \cdot b)|$.
\STATE \textbf{return} $(g = 1, d = d^\ast, \tau = \tau^\ast)$.
\end{algorithmic}
\end{algorithm}

The foreign threshold $F^\ast$ is calibrated per machine as the
$75$th percentile of $F$ over training-set write events, producing
roughly $25\%$ of writes triggering non-trivial foreign discount.

The target depth function $D^\ast(s, b, r)$ interpolates the substrate
retrieval curve to determine what accumulated depth is needed
for a fact of stability $s$ and base decay $b$ to remain retrievable
through its expected query horizon. Reinforcement events use a
slightly higher $D^\ast$ than first-write events to preserve
accumulation opportunity for future queries.

\subsection{State Features}
\label{app:features}

The controller conditions on six feature families summarized in
Section~\ref{sec:features}. Full details:

\paragraph{Dense semantic embedding.} A projection of the fact
template's last-token representation from the base model, passed
through an input normalization and a small MLP (\texttt{sem\_proj})
that decorrelates raw LM features from lexical surface form.
Dimension $256$.

\paragraph{GRU hidden state.} A recurrent summary of the trajectory
so far, updated per event. Takes the current fact's dense embedding,
event type indicator, and time delta since the previous event.
Provides temporal context that a bag-of-features representation
would lose. Dimension $256$.

\paragraph{Rank ages.} For each of the $R$ rank columns, the time
in days since that column was last written. Long-inactive columns
are candidates for aggressive writes because their existing content
has already decayed; recently touched columns carry current content
that will be interfered with. Dimension $R$.

\paragraph{Slot norm.} The $L_2$ norm of the mask-restricted LoRA
weights, either measured directly on the substrate (during online
training and evaluation) or predicted from ledger state via the
fitted regression from Appendix~\ref{app:physics} (during offline
training, for a $15\times$ speedup). Dimension $1$.

\paragraph{Predicted stability.} A scalar prediction from a dedicated
head trained via MSE against the ground-truth stability score.
Enables the controller to condition depth and gate decisions on
predicted fact durability without needing symbolic access to metadata
at inference.

\paragraph{Mask-level substrate features and forward-pass signal.} For the specific mask
that would receive this write, mean rank age within the mask, mean
foreign-step count within the mask, and the forward pass test
hit (Appendix~\ref{app:physics}). The forward pass provides a substrate-derived signal of whether the incoming value is already represented in the current slot, complementing the rank-age and interference features.AUC against
ground-truth reinforcement labels reaches 0.99. Dimension
$N_{\text{rank feats}} + 1$.

\subsection{Tau Head}
\label{app:tau}

The $\tau$ head is a five-way classifier over $\{25, 100, 400, 1600,
20000\}$ days. For a fact with base decay $b$, the target lifespan class is

\[ \tau^\ast(b)=\arg\min_{\tau_j\in\mathcal{T}} \left|\log \tau_j-\log(24.5\,b)\right|, \]

where $\mathcal{T}=\{25,100,400,1600,20000\}$ days.

\paragraph{Class weighting.} The $\tau$ head uses inverse-frequency
class weights since the
target distribution is more skewed. Bin $0$ ($\tau = 25$) receives
roughly $46\%$ of events, bin $1$ ($\tau = 100$) receives $10\%$,
bin $3$ ($\tau = 1600$) receives $2\%$. Inverse-frequency
weighting maintains recall on rare bins at the cost of noisier
training signal.

\subsection{Empirical Bias Initialization}
\label{app:bias_init}

Each action head's output bias is initialized to $\log(f)$ where $f$
is the marginal frequency of each action in the expert policy on a
calibration set of $150$ users. Table~\ref{tab:bias_init} shows the
initialized values.

\begin{table}[h]
\centering
\small
\begin{tabular}{ll}
\toprule
Head & Initialized bias values \\
\midrule
Store & $0.99$ (empirical write rate $72\%$) \\
Depth & $[-0.67, -4.47, -3.77, -2.01, -2.23, -2.03, -2.52]$ over $[20, 30, 40, 60, 80, 110, 150]$ \\
$\tau$ & $[-0.77, -2.26, -1.35, -3.81, -1.88]$ over $[25, 100, 400, 1600, 20000]$ \\
\bottomrule
\end{tabular}
\caption{Empirical bias initialization from expert marginals on
$150$-user calibration set.}
\label{tab:bias_init}
\end{table}

The depth bias is heavily negative on rare bins (e.g., $d = 30$ at
$-4.47$ reflects that only $1\%$ of expert writes use this depth).
The controller learns to adjust these biases per fact context and the
initialization ensures early training does not have to first recover
the marginal distribution.

\subsection{Lifecycle-Class Depth Prior}
\label{app:pattern_prior}

The prior matrix $W \in \mathbb{R}^{5 \times 7}$ maps predicted lifecycle-class probabilities to depth-preference adjustments. 

\begin{table}[h]
\centering
\small
\begin{tabular}{lccccccc}
\toprule
Pattern & $d = 20$ & $d = 30$ & $d = 40$ & $d = 60$ & $d = 80$ & $d = 110$ & $d = 150$ \\
\midrule
STABLE    & $-0.46$ & $-0.46$ & $-0.46$ & $-0.42$ & $-0.19$ & $+0.99$ & $+0.99$ \\
ROUTINE   & $+1.15$ & $+0.99$ & $+0.60$ & $-0.30$ & $-0.74$ & $-0.85$ & $-0.85$ \\
CONTESTED & $+1.03$ & $+1.03$ & $+0.74$ & $-0.18$ & $-0.75$ & $-0.93$ & $-0.93$ \\
TRANSIENT & $+1.15$ & $+0.99$ & $+0.60$ & $-0.30$ & $-0.74$ & $-0.85$ & $-0.85$ \\
ONESHOT & $-0.60$ & $-0.31$ & $+0.11$ & $+0.98$ & $+0.98$ & $-0.31$ & $-0.85$ \\
\bottomrule
\end{tabular}
\caption{Pattern-conditional depth prior matrix $W$. Each row is
mean-centered so the prior does not shift aggregate write rate. The
prior is combined with the pattern head's softmax as
$\alpha_{\text{pattern}} \cdot p_{\text{pattern}} \cdot W$ and added
to the depth head's raw logits.}
\label{tab:pattern_prior}
\end{table}

A parallel stability-conditional depth prior encourages the depth head
toward deep bins when predicted stability is high, using a
seven-dimensional vector $[-1, -1, -0.5, 0, 0.5, 1.5, 1.5]$ over the
depth bins, scaled by $\alpha_{\text{stab}}$ and the predicted
stability scalar. 

\subsection{Per-Router Hyperparameters}
\label{app:hyper}

Building on the bias initialization (Appendix~\ref{app:bias_init})
and depth priors (Appendix~\ref{app:pattern_prior}) above, this
section gives the remaining per-model hyperparameters. The prior
scales differ between routers due to different substrate retrieval curves.

\begin{table}[h]
\centering
\small
\begin{tabular}{lcc}
\toprule
Hyperparameter & PMM-PR & PMM-LR \\
\midrule
$\alpha_{\text{pattern}}$ & $1.0$ & $0.5$ \\
$\alpha_{\text{stab}}$ & $2.5$ & $0.5$ \\
$w_{\tau}$ base weight & $1.0$ & $1.0$ \\
$w_{\tau}^{\text{boost}}$ & $4.0$ & $4.0$ \\
$\tau$ boost duration (steps) & $200$ & $200$ \\
$\tau$ class weighting & inverse & inverse \\
Depth class weighting & sqrt & sqrt \\
DAgger $\beta$ schedule & $0.95 \rightarrow 0.50$ & $0.95 \rightarrow 0.50$ \\
Controller warmup (steps) & $100$ & $100$ \\
Trunk learning rate & $3 \times 10^{-4}$ & $3 \times 10^{-4}$ \\
Controller learning rate & $9 \times 10^{-4}$ & $9 \times 10^{-4}$ \\
BC epochs & $2$ & $2$ \\
Batch size (users per step) & $2$ & $2$ \\
\bottomrule
\end{tabular}
\caption{Hyperparameters for the two model configurations. PMM-LR uses
smaller prior scales because dynamic routing's permissive shallow-bin
dynamics would otherwise push the controller to over-invest in deep
writes.}
\label{tab:hyper}
\end{table}

\subsection{Stability Head Pretraining}
\label{app:stagea}

The stability head is pretrained via MSE regression against ground-truth stability scores over template-stability pairs from the training user pool. On $920$ unseen templates (held-out from $34$ users not seen during training), Stage A pretraining reaches $R^2 = 0.22$ at epoch $10$, peaking slightly before plateauing. 

The template-only prediction ceiling (best any model could achieve from the template embedding alone) is approximately $0.36$. The gap from 0.22 to 0.36 is due to residual variance the head cannot extract from the embedding alone. Meanwhile, the gap from 0.36 to 1.00 represents the limit of this feature as identical fact templates have different true stabilities across different users depending on their personal patterns. This limit flows downstream to the $\tau$ head, which reads from the stability prediction.

\subsection{STABLE Analysis Details}
\label{app:stable}

Per-query diagnostics on STABLE VAL outcomes were collected on $56$
STABLE VAL queries across a $5$-user analysis sample using the PMM-LR
controller.
\begin{figure*}[t]
\centering
\begin{tikzpicture}[
    node distance=1.0cm and 1.2cm,
    box/.style={rectangle, draw=black!80, thick, fill=blue!5, text width=3.3cm, align=center, rounded corners=1mm, minimum height=1.3cm, font=\footnotesize},
    data/.style={cylinder, draw=black!80, thick, aspect=0.3, fill=green!10, text width=3.2cm, align=center, minimum height=1.3cm, shape border rotate=90, font=\footnotesize},
    arrow/.style={->, thick, >=Stealth, draw=black!80}
]

\node[box, fill=purple!10] (llm) {\textbf{Phase 1: LLM Gen.}\\Motif Sampling \& Masking};
\node[data, right=of llm] (raw) {Raw Masked Ledgers};
\node[box, right=of raw] (pattern) {\textbf{LLM Pruning}};

\node[box, below=of pattern, yshift=-0.2cm] (physics) {\textbf{Storage}\\$Decay_{eff} = D_0 \times 1.8^{streak}$};
\node[box, left=of physics] (expand) {\textbf{Trajectory Expansion}\\Constructive Accumulation};
\node[box, left=of expand] (queries) {\textbf{Query Injector}\\Placement via Decay Curve};

\node[data, below=of queries, yshift=-0.5cm] (final) {Final Dataset};

\draw[arrow] (llm) -- (raw);
\draw[arrow] (raw) -- (pattern);

\draw[arrow] (pattern) -- (physics);

\draw[arrow] (physics) -- (expand);
\draw[arrow] (expand) -- (queries);

\draw[arrow, dashed] (queries) -- 
    node[left, align=right, font=\scriptsize, xshift=-0.1cm] {Fresh: $t < D_{eff}$\\$\implies$ \textbf{VALUE Target}} 
    node[right, align=left, font=\scriptsize, xshift=0.1cm] {Stale: $t > D_{eff}$\\$\implies$ \textbf{ABSTAIN Target}} 
(final);

\begin{scope}[on background layer]
    \fill[gray!10, rounded corners] 
        ([xshift=-0.3cm, yshift=0.4cm]llm.north west) rectangle 
        ([xshift=0.3cm, yshift=-0.3cm]raw.south east);
    \node[anchor=north west, font=\itshape\scriptsize, color=gray!80] at ([xshift=-0.2cm, yshift=0.3cm]llm.north west) {Semantic Generation};

    \fill[orange!5, rounded corners] 
        ([xshift=-0.3cm, yshift=0.4cm]pattern.north west) rectangle 
        ([xshift=0.3cm, yshift=-0.3cm]queries.south east);
    \node[anchor=north east, font=\itshape\scriptsize, color=orange!80] at ([xshift=0.2cm, yshift=0.3cm]pattern.north east) {Physics \& Enforcement};
\end{scope}

\end{tikzpicture}
\caption{Overall dataset generation pipeline. Phase 1 grounds the data via LLM generation, while Phase 2 strictly enforces temporal dynamics, constructive accumulation, and delay-calibrated query injection.}
\label{fig:pipeline}
\end{figure*}
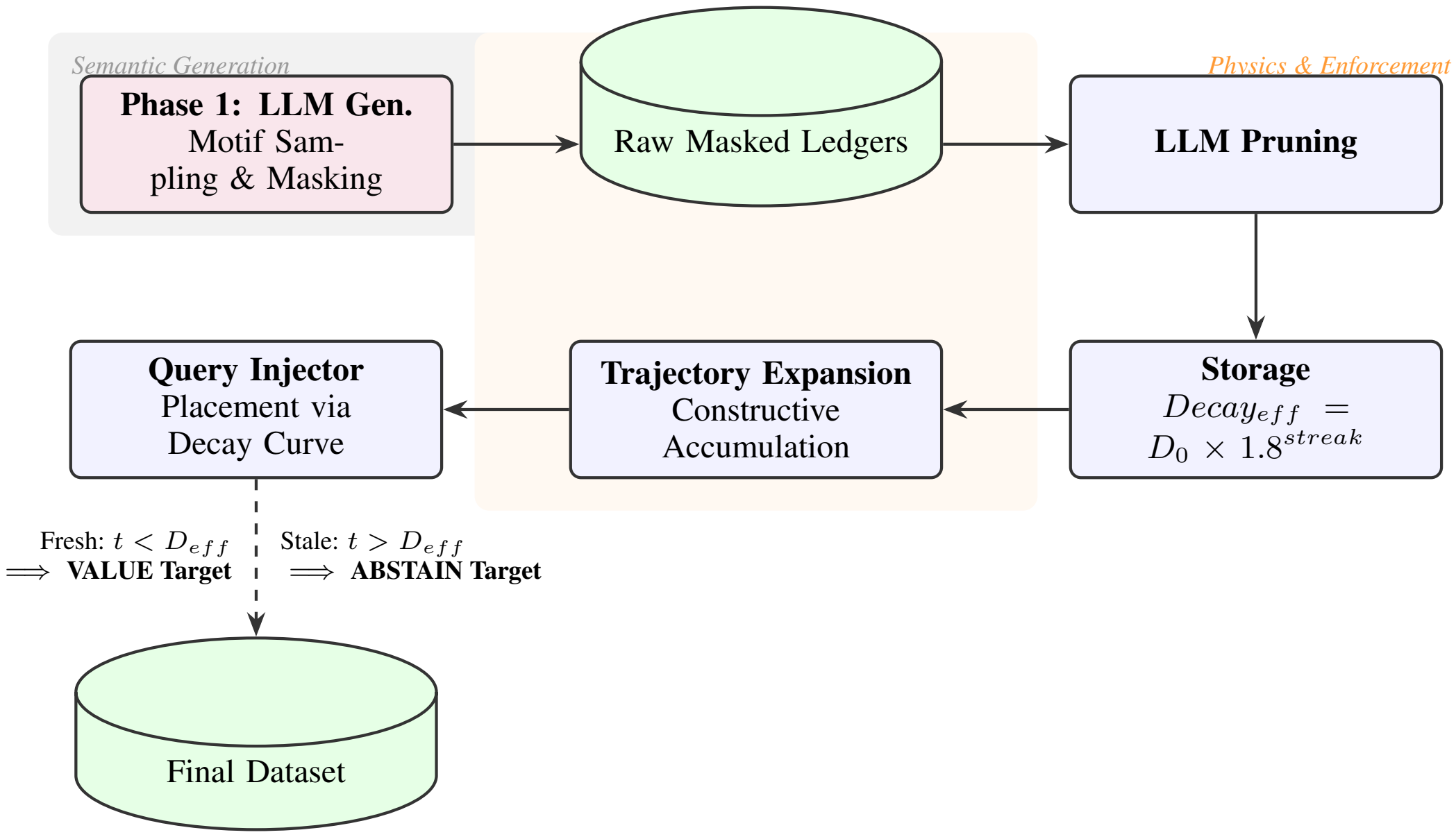

\begin{table}[h]
\centering
\small
\begin{tabular}{lcc}
\toprule
& Success ($n = 28$) & Failure ($n = 28$) \\
\midrule
Prior writes (mean) & $1.39$ & $1.11$ \\
Last write depth (mean) & $109.6$ & $104.3$ \\
Accumulated depth (mean) & $142.1$ & $113.9$ \\
Days since last write (mean) & $18.5$ & $536.7$ \\
Self steps at query & $124.1$ & $57.2$ \\
Foreign steps at query & $1127.8$ & $1205.4$ \\
Foreign-to-self ratio (median) & $8.6$ & $23.2$ \\
\bottomrule
\end{tabular}
\caption{Per-query diagnostics on STABLE VAL. Successes and failures
similar on writing behavior (depth, prior writes) but differ
dramatically on query horizon and foreign-to-self ratio at query
time.}
\label{tab:stable_diag}
\end{table}

The distinguishing features between success and failure are query
horizon and foreign-to-self ratio at query time. Successful STABLE
queries fire on average $18.5$ days after last write; failed queries
fire $536.7$ days after. Failed queries suffered from a $2.7\times$ higher foreign-to-self gradient ratio. This was not caused by encountering more absolute foreign traffic (means 1205 vs. 1128), but by the stable fact receiving far fewer reinforcing writes to defend its own trace against that traffic (57 vs. 124).

\paragraph{Max-$\tau$ ablation.} An alternative $\tau$ update rule
that preserves the longest $\tau$ any fact has assigned to a column
(rather than overwriting) protects STABLE facts on shared columns
from subsequent shorter-$\tau$ writes. This variant achieves
class-balanced accuracy $0.45$ ($5$-user eval), with STABLE VAL
improving by $\sim 13$ points but ABS accuracy collapsing across
TRANSIENT, ROUTINE, CONTESTED, and ONESHOT (each below $0.05$).
TRANSIENT facts written to shared columns accidentally inherit the long $\tau$ of older STABLE facts. Consequently, these volatile facts never decay, causing the system to provide outdated answers instead of correctly abstaining during stale queries. This shows that the trade-off cannot be resolved by this simple max-\(\tau\) update rule and reflects the coupling introduced by shared-column compression.

\paragraph{Reinforcement effect.} STABLE facts written multiple
times over their trajectory re-assert long $\tau$ on their columns
and refresh self-investment against accumulated foreign traffic.
STABLE facts with $3+$ writes hit at $0.68$ versus $0.51$ for
single-write STABLE. The controller does not independently trigger
reinforcement and it is driven by the upstream event stream
identifying repeated observations of the same fact.

\subsection{Phase 2 PPO Details}
\label{app:phase2}

Phase 2 refines the phase 1 controller via PPO. Reward is only outcome-based for correct VAL and ABS queries. The
phase 1 checkpoint serves as the reference policy for KL
regularization.

 KL divergence to reference is penalized with weight
$0.10$ on the store head and $0.02$ on depth. Batch size is $4$ users
per PPO iteration, with $3$ PPO epochs and minibatch size $32$. Value
coefficient $0.5$, auxiliary loss coefficient $0.3$, entropy
coefficient $0.015$ on depth and $0.005$ on the store head. 

In our experiments, phase 2 provided marginal improvement over phase
$1$ (roughly $\pm 1$ point class-balanced), primarily by shifting
the policy's VAL-ABS balance without meaningfully improving aggregate.
We attribute this to phase 1 already being close to the metadata oracle
(${\sim}86\%$), leaving little room for refinement via reward signal.
We treat phase 2 as an optional stage.

\subsection{Input Feature Ablations}
\label{app:ablations}

To understand which information is strictly necessary for parametric memory management, we performed a set of input feature ablations. Because the controller uses a shared representation trunk that feeds multiple heads, simply removing an architectural head could confound the results. Instead, we ablate by zeroing out the specific input sources entirely before they reach the controller, directly testing the information bottleneck. We evaluate four ablations against the PMM-PR baseline:
\begin{enumerate}[label=\arabic*., leftmargin=*]
    \item \textbf{No Semantic:} Zeroes out the dense text embedding.
    \item \textbf{No Priors:} Removes the Stage A stability warm-start and the mathematical depth-bias.
    \item \textbf{No History:} Zeroes out the GRU trajectory hidden state.
    \item \textbf{No Substrate:} Zeroes out substrate readings (\texttt{probe\_hit}, \texttt{slot\_norm}, \texttt{foreign\_steps}).
\end{enumerate}

\begin{table}[h]
\centering
\scriptsize
\begin{tabular}{lccccccccccc}
\toprule
& & \multicolumn{2}{c}{STABLE} & \multicolumn{2}{c}{ROUTINE} & \multicolumn{2}{c}{CONTESTED} & \multicolumn{2}{c}{TRANSIENT} & \multicolumn{2}{c}{ONESHOT} \\
\cmidrule(lr){3-4} \cmidrule(lr){5-6} \cmidrule(lr){7-8} \cmidrule(lr){9-10} \cmidrule(lr){11-12}
System & CB & VAL & ABS & VAL & ABS & VAL & ABS & VAL & ABS & VAL & ABS \\
\midrule
PMM-PR & \textbf{0.60} & 0.33 & 0.81 & 0.80 & 0.64 & 0.81 & 0.34 & 0.66 & 0.58 & 0.82 & 0.58 \\
\midrule
No History & 0.58 & 0.33 & 0.91 & 0.67 & 0.68 & 0.67 & 0.32 & 0.74 & 0.68 & 0.58 & 0.76 \\
No Substrate & 0.58 & 0.34 & 0.95 & 0.77 & 0.55 & 0.76 & 0.28 & 0.70 & 0.55 & 0.65 & 0.72 \\
No Priors & 0.52 & 0.42 & 0.80 & 0.66 & 0.57 & 0.73 & 0.20 & 0.78 & 0.41 & 0.69 & 0.44 \\
No Semantic & 0.47 & 0.28 & 0.88 & 0.70 & 0.33 & 0.63 & 0.28 & 0.61 & 0.35 & 0.49 & 0.60 \\
\bottomrule
\end{tabular}
\caption{Input feature ablations. Losing semantic features or priors causes substantial failure. History and Substrate act as a dual-redundancy system, but isolating them reveals specific, severe capability collapses (e.g., losing history crashes ONESHOT VAL by 24 points).}
\label{tab:ablations}
\end{table}

The per-class VAL/ABS splits (Table~\ref{tab:ablations}) reveal exactly how the controller uses its inputs, surfacing three core findings:

\paragraph{1. Semantic embeddings and priors provide substantial management signal.}
Without semantic embeddings, the model cannot map an incoming fact to a behavioral pattern. Without this information, the policy falls back toward less differentiated write behavior, reducing its ability to separate persistent and volatile events. Removing semantic features lowers CB to \(0.47\), while removing the substrate-derived priors lowers it to \(0.52\).

\paragraph{2. History and substrate state provide complementary information.}
The \emph{No History} (CB 0.58) and \emph{No Substrate} (CB 0.58) ablations survive in aggregate because the GRU and the LoRA state provide overlapping signals for reinforcement. If the GRU is removed, the substrate's \texttt{probe\_hit} still tells the controller a fact is recurring. If the substrate is removed, the GRU's temporal memory knows it has seen the embedding multiple times. However, the class-wise splits show they are not perfectly redundant:
\begin{itemize}[leftmargin=*]
    \item \textbf{History disambiguates lifecycle from trajectory context:} Without the GRU, the model is blind to the timeline. When a ONESHOT fact arrives, the model cannot confidently assign it moderate depth because it lacks the behavioral context to separate it from the first write of a ROUTINE fact. Consequently, it hedges its bets, and ONESHOT VAL crashes by $24$ points ($0.82 \rightarrow 0.58$).
    \item \textbf{Substrate state exposes the consequences of prior writes:} Without the substrate, the model cannot see collision metrics (\texttt{foreign\_steps}, \texttt{slot\_norm}). Blind to overlap, it overwrites frequently. This leaves stale traces persisting when they should have decayed, dropping abstention (ABS) accuracy on volatile, highly-reinforced classes (ROUTINE ABS drops $0.64 \rightarrow 0.55$; CONTESTED ABS drops $0.34 \rightarrow 0.28$).
\end{itemize}

\paragraph{3. Higher abstention under ablation reflects more conservative writing.}
STABLE and ONESHOT ABS actually improve under these ablations (climbing to 0.91-0.95), but this is not a genuine gain as the controller compensates for information loss by defaulting to shallower writes, which decay faster and inflate ABS at the cost of VAL.

\section{Dataset and Trajectory Generation Details}
\label{app:dataset}

This appendix provides extended statistics, motif distributions, and qualitative examples for the controlled long-horizon user trajectories introduced in Section~2.

\subsection{Dataset Statistics}

\begin{figure}[h]
    \centering
    \includegraphics[width=0.45\textwidth]{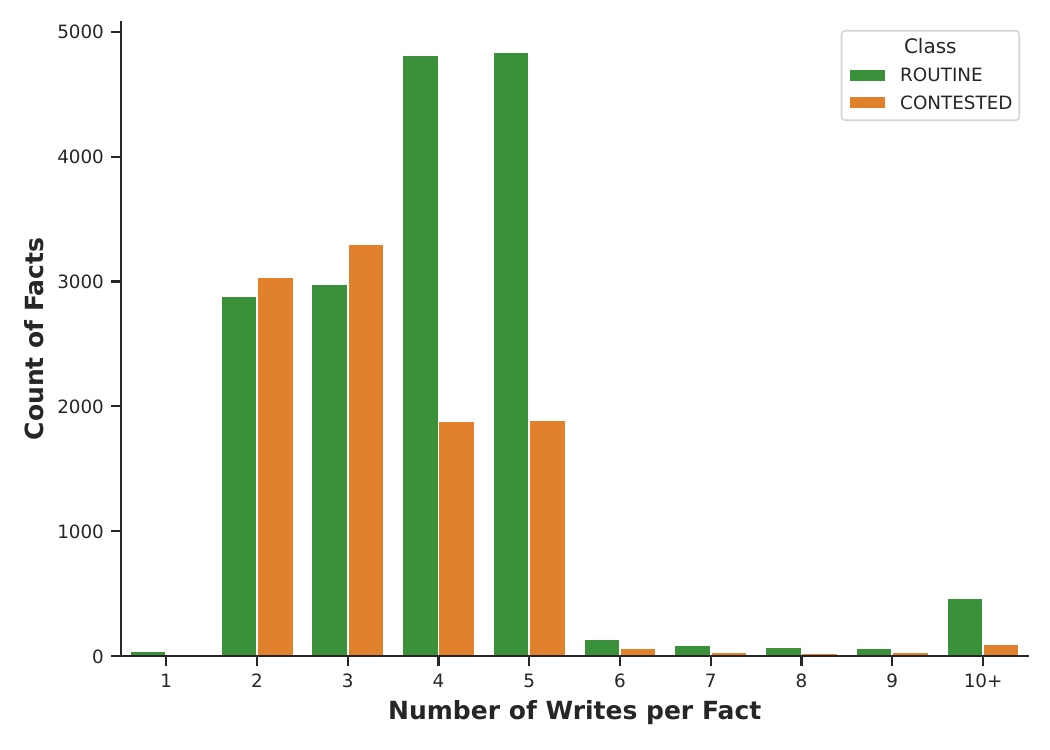} 
    \caption{Two facts can have very similar properties such as stability and number of writes, yet belong to very different classes based on user history and semantic meaning.}
    \label{fig:empirical_dist}
\end{figure}

\begin{figure}[h]
    \centering
    \includegraphics[width=0.6\textwidth]{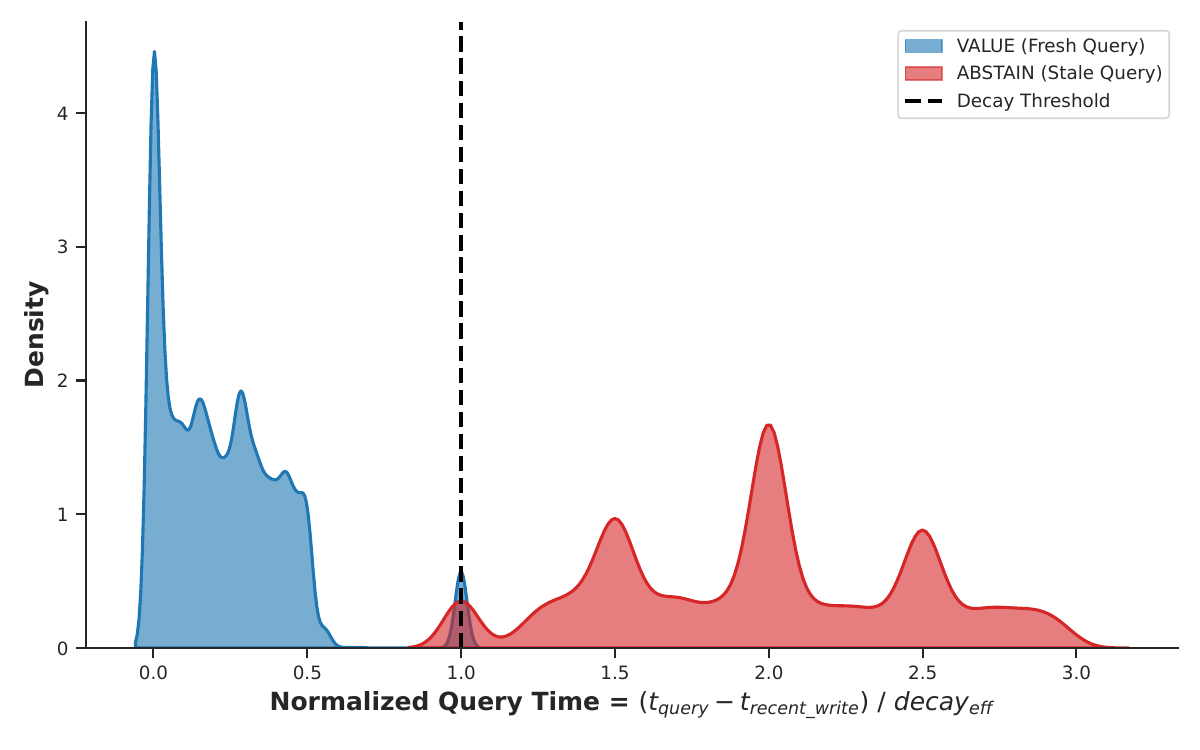} 
    \caption{Fresh versus stale query injection density. Retrieval (VAL) and abstention (ABS) queries are placed relative to each fact's validity horizon.}
    \label{fig:fresh_stale}
\end{figure}

The dataset consists of $909$ user trajectories spanning multi-year horizons. Table~\ref{tab:dataset_stats} summarizes the aggregate dataset statistics and the class balance across the five memory classes. Figure~\ref{fig:empirical_dist} visualizes how identical fact properties map to different classes, while Figure~\ref{fig:fresh_stale} illustrates the query injection density across a fact's decay threshold.

\begin{table}[h]
\centering
\small
\begin{tabular}{ll}
\toprule
\textbf{Metric} & \textbf{Value} \\
\midrule
Total Users & $909$ \\
Query Horizon & Avg $5,434$ days (Min: $309$, Max: $25,282$) \\
Total Writes & $170,121$ \\
Total Queries & $144,164$ \\
Query Target Split & $69.3\%$ VALUE $\mid$ $30.7\%$ ABSTAIN \\
\bottomrule
\end{tabular}
\vspace{1em}

\begin{tabular}{lcc}
\toprule
\textbf{Classes} & \textbf{\% of Writes} & \textbf{\% of Queries} \\
\midrule
STABLE & $27.2\%$ & $46.7\%$ \\
ROUTINE & $41.0\%$ & $24.3\%$ \\
CONTESTED & $20.5\%$ & $11.4\%$ \\
TRANSIENT & $7.7\%$ & $8.9\%$ \\
ONESHOT & $3.7\%$ & $8.6\%$ \\
\bottomrule
\end{tabular}
\caption{At-a-glance dataset statistics (top) and class balance distributions (bottom). Queries are intentionally reweighted to ensure critical behaviors like expiration (TRANSIENT) are adequately represented despite write imbalances. Query horizon is separate from write events horizon of 700-1500 days.}
\label{tab:dataset_stats}
\end{table}


\subsection{Motif Distribution and Cross-User Variance}

\begin{figure}[h]
    \centering
    \includegraphics[width=0.7\textwidth]{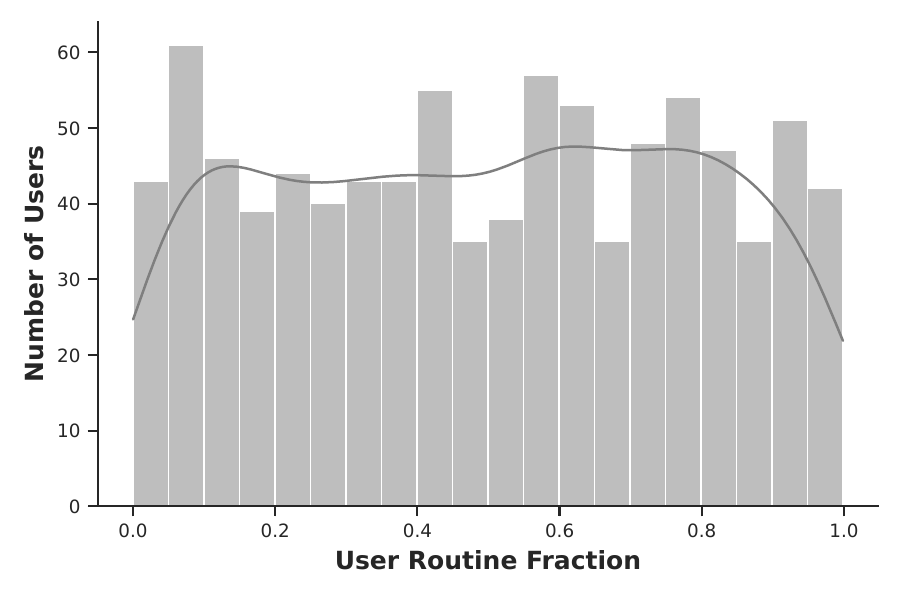} 
    \caption{Cross-user variance in routine prevalence. The variation across users illustrates why lifespans cannot be determined solely from fact type and must also depend on individual trajectory context.}
    \label{fig:cross_user}
\end{figure}

To ensure heterogeneous, user-level memory dynamics, user personas were generated using a variety of underlying temporal motifs governing their behavior (Table~\ref{tab:motifs}). Because these motifs dictate behavior at the user level, the same fact type (e.g., a hobby or a job) can be stable for one user but volatile for another. Figure~\ref{fig:cross_user} illustrates the resulting cross-user variance in routine predictability.

\begin{table}[h]
\centering
\small
\begin{tabular}{lc}
\toprule
\textbf{Temporal Motif} & \textbf{Users} \\
\midrule
NORMAL DRIFT: Slow, mundane, almost imperceptible changes & $347$ \\
THE MICRO-VARIATION: Highly stable daily routine & $75$ \\
THE SEASONAL SHIFT: 2-to-3 month block behavior & $67$ \\
THE PHASED TRANSITION: Gradual shift over 2 months & $59$ \\
THE VACATION/TRIP: Stable routine changes for 2 weeks & $51$ \\
THE INSTANT CORRECTION: Realizes mistake minutes/hours later & $45$ \\
THE CASCADE EVENT: Single major event triggers 3-4 routines & $41$ \\
THE FAILED HABIT: Tries for 5 days, then drops completely & $36$ \\
THE CONTEXT-DEPENDENT TOGGLE: Toggles based on day of week & $33$ \\
THE RELAPSE: Abandoned habit returns heavily for 2 weeks & $30$ \\
DORMANT RESURGENCE: Ignored for 6 months, then relevant & $26$ \\
THE RAPID PROMOTER: Switches roles every 3-4 months & $24$ \\
THE VOLATILE ROUTINE: Constantly breaks habit and gets back & $22$ \\
THE NOMAD: Moves and changes routine every 6 months & $20$ \\
THE CHAMELEON: Adopts completely new hobbies every season & $18$ \\
THE BINGE: Intense, short-term burst never happening again & $15$ \\
\bottomrule
\end{tabular}
\caption{Distribution of the underlying temporal motifs used to govern user memory dynamics during dataset generation.}
\label{tab:motifs}
\end{table}

\subsection{Qualitative Example}

Table~\ref{tab:qualitative} provides a qualitative snippet of a user trajectory demonstrating how writes, updates, and queries are interleaved. Entities are masked to decouple semantic world knowledge from parametric retention. 

\begin{table}[h]
\centering
\small
\begin{tabular}{llll}
\toprule
\textbf{Day} & \textbf{Event} & \textbf{Pattern} & \textbf{Ledger Entry / Query} \\
\midrule
Day 0 & INTRODUCE & STABLE & I work as a [JOB\_8] \\
Day 2 & INTRODUCE & ROUTINE & I ordered my morning coffee from [LOC\_CAFE\_4] \\
Day 5 & INTRODUCE & TRANSIENT & I grabbed a quick lunch with Sarah at [LOC\_BISTRO\_9] \\
Day 12 & ABS QUERY & TRANSIENT & I grabbed a quick lunch with Sarah at [MASK] \\
Day 14 & INTRODUCE & CONTESTED & My current priority project is [PROJ\_2] \\
Day 15 & UPDATE & ROUTINE & I ordered my morning coffee from [LOC\_CAFE\_4] \\
Day 18 & VAL QUERY & ROUTINE & I ordered my morning coffee from [MASK] \\
Day 22 & CONTRADICT & CONTESTED & My current priority project is [PROJ\_7] \\
Day 25 & VAL QUERY & CONTESTED & My current priority project is [MASK] \\
Day 30 & INTRODUCE & ONESHOT & I booked a 2-week vacation to [LOC\_1] \\
Day 45 & VAL QUERY & ONESHOT & I booked a 2-week vacation to [MASK] \\
Day 400 & VAL QUERY & STABLE & I work as a [MASK] \\
\bottomrule
\end{tabular}
\caption{A narrative snippet of a user trajectory demonstrating all five memory lifecycles. The example illustrates the distinct target behaviors: TRANSIENT information is queried after its validity horizon, CONTESTED information is queried after a value revision, and STABLE information must remain retrievable across long intervening histories. The non-query inputs also follow the same [MASK]-Output format to memorize as the query inputs but are presented in a readable form here with outputs values inserted in their respective [MASK] positions.}
\label{tab:qualitative}
\end{table}

\subsection{Benchmark Comparison Criteria}
\label{app:benchmark_comparison}

Table~\ref{tab:benchmark-comparison} compares prior long-term memory benchmarks along the following properties:

\paragraph{Belief revision.}
Whether later observations can supersede an earlier value and evaluation requires recovering the currently valid state.

\paragraph{Time-based expiry.}
Whether previously valid information can become invalid solely because its validity horizon has elapsed, without an explicit replacement, correction, or deletion event.

\paragraph{Reinforcement.}
Whether repeated consistent observations affect subsequent memory behavior, such as strengthening retention, rather than being treated as independent events.

\paragraph{User-conditioned lifespan.}
Whether the appropriate lifespan of a memory depends on the individual user's observed history rather than only on the semantic type of the information.

\paragraph{Lifecycle supervision.}
Whether the benchmark provides event-level ground truth describing how a memory's validity or state evolves over time.

We mark a property as \emph{full} when it is explicitly constructed and evaluated, \emph{partial} when related behavior is present but not directly isolated, and \emph{absent} otherwise.

\subsection{Dataset Schema and Comprehensive Trajectory Example}
\label{app:dataset_schema}

Because the full dataset consists of multi-year trajectories with hundreds of events per user, we provide a detailed example here to illustrate the exact inputs, metadata, and query formats utilized by both our parametric controllers and the retrieval-based baselines.

Each user in the dataset is represented as a JSON object containing a \texttt{facts\_metadata} dictionary (the ground-truth for each fact) and an \texttt{events} array (the chronological stream of writes and queries). 

\paragraph{1. Ground-Truth Metadata Dictionary}
The \texttt{facts\_metadata} dictionary maps each semantic fact type (e.g., \textit{favorite\_coffee\_shop}) to its ground-truth lifecycle properties which represents a user's belief states. This metadata is strictly hidden from the controller at inference time but is used by the DAgger oracle during Phase 1 training to compute target depths.

\begin{verbatim}
"facts_metadata": {
  "user_profession": {
    "write_pattern": "STABLE",
    "stability_score": 0.92,
    "base_decay_days": 3650.0
  },
  "morning_coffee_spot": {
    "write_pattern": "ROUTINE",
    "stability_score": 0.45,
    "base_decay_days": 14.0
  },
  "current_project": {
    "write_pattern": "CONTESTED",
    "stability_score": 0.65,
    "base_decay_days": 90.0
  },
  "todays_lunch": {
    "write_pattern": "TRANSIENT",
    "stability_score": 0.12,
    "base_decay_days": 2.0
  }
}
\end{verbatim}

\paragraph{2. Chronological Event Stream}
The \texttt{events} array represents the timeline. Each event contains a timestamp, the event type (\texttt{INTRODUCE}, \texttt{UPDATE}, \texttt{CONTRADICT}, or \texttt{QUERY}), and the event. 

Below is an illustrative sequence showing how the \texttt{morning\_coffee\_spot} (a ROUTINE fact) and \texttt{todays\_lunch} (a TRANSIENT fact) are processed over a 20-day window.

\vspace{1em}
\noindent \textbf{Event 1: First observation of a routine (Day 0)}
\begin{verbatim}
{
  "clock_days": 0.0,
  "event_type": "INTRODUCE",
  "fact_type": "morning_coffee_spot",
  "value": "CAFE_4",
  "is_reinforcement": false,
  "template_input": "I ordered my morning coffee from [MASK]",
  "natural_language": "[Clock: 0000] I ordered my morning coffee 
  from CAFE_4."
}
\end{verbatim}

\noindent \textbf{Event 2: A transient, unrepeated event (Day 2)}
\begin{verbatim}
{
  "clock_days": 2.5,
  "event_type": "INTRODUCE",
  "fact_type": "todays_lunch",
  "value": "RESTAURANT_9",
  "is_reinforcement": false,
  "template_input": "I grabbed a quick lunch with Sarah at [MASK]",
  "natural_language": "[Clock: 0002] I grabbed a quick lunch with 
  Sarah at RESTAURANT_9."
}
\end{verbatim}

\noindent \textbf{Event 3: Reinforcement of the routine (Day 5)}
\begin{verbatim}
{
  "clock_days": 5.0,
  "event_type": "UPDATE",
  "fact_type": "morning_coffee_spot",
  "value": "CAFE_4",
  "is_reinforcement": true,
  "template_input": "I ordered my morning coffee from [MASK]",
  "natural_language": "[Clock: 0005] I ordered my morning coffee 
  from CAFE_4."
}
\end{verbatim}

\noindent \textbf{Event 4: Stale Abstention Query for the Transient fact (Day 15)} \\
\textit{Because \texttt{todays\_lunch} has a base decay of 2.0 days, and 12.5 days have passed since Event 2, the ground-truth target strictly demands abstention.}
\begin{verbatim}
{
  "clock_days": 15.0,
  "event_type": "QUERY",
  "target_type": "ABSTAIN",
  "fact_type": "todays_lunch",
  "target_value": "Unknown",
  "template_input": "I grabbed a quick lunch with Sarah at [MASK]",
  "natural_language": "Where did you grab lunch with Sarah?"
}
\end{verbatim}

\noindent \textbf{Event 5: Value Retrieval Query for the Routine fact (Day 18)} \\
\textit{Because the routine was reinforced at Day 5, the controller should have accumulated depth, keeping the memory valid for this query 13 days later.}
\begin{verbatim}
{
  "clock_days": 18.0,
  "event_type": "QUERY",
  "target_type": "VALUE",
  "fact_type": "morning_coffee_spot",
  "target_value": "CAFE_4",
  "template_input": "I ordered my morning coffee from [MASK]",
  "natural_language": "Question: Where do you usually
  order your morning coffee?"
}
\end{verbatim}

\section{\new{Natural-Language Memory Management End to End}}
\label{app:nl_study}

\begingroup\newcolor
The main paper isolates memory management with a templated interface. Here we run the \emph{complete} system with natural-language (NL) writes and queries: the same substrate, controller architecture, action space and training recipe (\S\ref{sec:controller}). Only quantities that depend on the input format are re-measured. On the same held-out users, NL performance is within $0.02$ class-balanced (CB) accuracy of templated performance for the oracle, PMM-PR and PMM-LR (Table~\ref{tab:main_results}).

\subsection{Setup}
Each write is expressed as $10$ question--answer pairs about the fact (e.g., \textit{Where did I take my dog?} $\rightarrow$ \textit{the park}). This means that natural language interfacing requires an additional question-answer generation module to ensure correct writes. Model editing approaches may further help with the write step and do away with the question generation module. A depth-$d$ write performs $d$ SGD steps on the routed columns, split evenly across the pairs, with the loss on answer tokens. Queries are free-form questions that do not repeat the write wording. The substrate mechanics are format-independent, but write strength and its survival under decay are not. Table~\ref{tab:nl_physics} measures both for NL writes. Two properties differ from the templated setting: recall plateaus early (reliable from $d\approx60$), and decay acts as a cliff rather than a gradual fade (recall is $0.88$--$0.96$ at $f=0.9$, but at most $0.19$ at $f=0.7$ and zero at $f=0.5$).

\endgroup

\begin{table}[h]
\newcolor
\centering
\small
\begin{tabular}{lcccc}
\toprule
& \multicolumn{4}{c}{decay factor $f=\exp(-\Delta t/\tau)$ at query time} \\
\cmidrule(lr){2-5}
depth $d$ & $1.0$ & $0.9$ & $0.7$ & $0.5$ \\
\midrule
$20$  & 0.23 & 0.12 & 0.00 & 0.00 \\
$40$  & 0.83 & 0.56 & 0.06 & 0.00 \\
$60$  & 0.96 & 0.88 & 0.08 & 0.00 \\
$110$ & 1.00 & 0.94 & 0.17 & 0.00 \\
$150$ & 1.00 & 0.96 & 0.19 & 0.00 \\
\bottomrule
\end{tabular}
\caption{Recall of a single NL fact vs.\ write depth and remaining decay factor (question--answer writes, matched retrieval adapter, $48$ facts).}
\label{tab:nl_physics}
\end{table}

\begingroup\newcolor
\subsection{Relation to Prior Work}
Extracting knowledge written into weights with free-form questions depends on how it was written. \citet{allenzhu2024physicslanguagemodels31} show that extractability requires diverse formulations of each fact during training, and \citet{mecklenburg2024injectingnewknowledgelarge} and \citet{ovadia-etal-2024-fine} find that fine-tuned question answering improves with paraphrase augmentation. Our lifecycle management (when to write, how deeply, for how long) transfers to natural language without change to the substrate, or controller training recipe.
\endgroup

\end{document}